\documentclass{IEEEtaes}

\usepackage{color}
\usepackage{array}
\usepackage{amsthm}
\usepackage{cite}
\usepackage{graphicx}
\usepackage{algorithm}
\usepackage{algpseudocode}
\usepackage{amsmath}
\usepackage{amssymb}
\usepackage{hyperref}
\usepackage{svg}
\usepackage{tabularray}
\usepackage{caption}
\usepackage{subcaption}

\jvol{XX}
\jnum{XX}
\jmonth{XXXXX}
\paper{1234567}
\pubyear{2024}
\doiinfo{TAES.2024.Doi Number}

\begin{document}

\makeatletter
\def\ps@plain{%
    \let\@mkboth\@gobbletwo
    \let\@oddhead\@empty
    \let\@evenhead\@empty
    \def\@oddfoot{\hfil\footnotesize \parbox{\textwidth}{This work has been submitted to the IEEE for possible publication.\newline Copyright may be transferred without notice, after which this version may no longer be accessible.}\hfil\thepage}
    \let\@evenfoot\@oddfoot
}
\pagestyle{plain}
\makeatother

\title{Aircraft Trajectory Segmentation-based Contrastive Coding: A Framework for Self-supervised Trajectory Representation}

\author{THAWEERATH PHISANNUPAWONG}
\author{JOSHUA J. DAMANIK}
\author{HAN-LIM CHOI} 
\member{Senior Member, IEEE}
\affil{Korea Advanced Institute of Science and Technology \newline Daejeon, 34141, Republic of Korea}

\receiveddate{
This work is supported by the Korea Agency for Infrastructure Technology Advancement (KAIA) grant funded by the Ministry of Land, Infrastructure and Transport (Grant 22DATM-C163373-02).}
\corresp{\itshape (Corresponding author: Han-Lim Choi)}

\authoraddress{Authors' addresses: Thaweerath Phisannupawong, Joshua J. Damanik, and Han-Lim Choi are with the Department of Aerospace Engineering, Korea Advanced Institute of Science and Technology, Daejeon, 34141, Republic of Korea, Email: (\href{mailto:petchthwr@kaist.ac.kr}{petchthwr@kaist.ac.kr}, \href{mailto:joshuad@kaist.ac.kr}{joshuad@kaist.ac.kr}, \href{mailto:hanlimc@kaist.ac.kr}{hanlimc@kaist.ac.kr})}

\supplementary{The aircraft trajectory classification datasets used in this paper are available at \href{https://huggingface.co/datasets/petchthwr/ATFMTraj}{huggingface.co/datasets/petchthwr/ATFMTraj}.\newline The source code is publicly available at \href{https://github.com/petchthwr/ATSCC}{github.com/petchthwr/ATSCC}.}

\maketitle

\begin{abstract}
Air traffic trajectory recognition has gained significant interest within the air traffic management community, particularly for fundamental tasks such as classification and clustering. This paper introduces Aircraft Trajectory Segmentation-based Contrastive Coding (ATSCC), a novel self-supervised time series representation learning framework designed to capture semantic information in air traffic trajectory data. The framework leverages the segmentable characteristic of trajectories and ensures consistency within the self-assigned segments. Intensive experiments were conducted on datasets from three different airports, totaling four datasets, comparing the learned representation's performance of downstream classification and clustering with other state-of-the-art representation learning techniques. The results show that ATSCC outperforms these methods by aligning with the labels defined by aeronautical procedures. ATSCC is adaptable to various airport configurations and scalable to incomplete trajectories. This research has expanded upon existing capabilities, achieving these improvements independently without predefined inputs such as airport configurations, maneuvering procedures, or labeled data.
\end{abstract}
\begin{IEEEkeywords} Air Traffic Management, Contrastive Learning, Representation Learning, Time Series, Trajectory Clustering
\end{IEEEkeywords}

\section{INTRODUCTION}
Air transportation plays a crucial role in the world economy; as economic growth increases, so does air traffic. Operational concepts have been developed to accommodate this continuous growth in air traffic, for the current air traffic management (ATM) system and procedures will not be able to maintain the level of safety and efficiency in handling the increasing traffic, especially for airports with dense traffic flow and complex maneuvering patterns \cite{TBO}. Therefore, airspace modernization has become a challenging aim for the ATM society and the regulatory authorities. The ATM concept is expected to shift from a clearance-based to a trajectory-based operation (TBO), which allows more flexible clearance and efficiency in utilizing airspace for high-density traffic. Although the trajectory is human-interpretable, managing higher density, more flexible traffic can be overwhelming for human operators; therefore, intelligence management and recognition systems have been developed to assist the air traffic controller in characterizing flight trajectories, estimating the airport capacity, and decision-making.

Nowadays, a vast amount of aircraft trajectory data, including Automatic Dependent Surveillance-Broadcast (ADS-B) recording, is publicly available. This availability holds significant potential for trajectory recognition research. Classification and clustering are frequently explored in literature as fundamental recognition tasks. For instance, classification assists air traffic controllers in organizing trajectories for monitoring and capacity estimation. Clustering has continuously attracted increased attention from the community, mainly because trajectory datasets are usually found unlabeled. However, the trajectories as time series are typically rich in information, complex, and highly dimensional. Many algorithms designed for these downstream tasks are vulnerable to the high dimensionality of data and usually necessitate feature extraction. Transforming them into a more generalizable data representation is known to be an effective method for enhancing downstream recognition task performance.

Self-supervised contrastive representation learning has shown effectiveness in enhancing recognition tasks in many applications, namely vision, natural language processing, and time series. Although this approach has been extensively applied to real-world time series data, including medical signals, electromagnetics, sound waves, and more, the exploration of moving object trajectories still offers considerable opportunity to be explored. This paper introduces Aircraft Trajectory Segmentation-based Contrastive Coding (ATSCC), a self-supervised contrastive representation learning framework designed for aircraft trajectory data motivated by the operational contextualization of ATM. An assumption has been made that the states within a trajectory segment share identical contexts. Consequently, ATSCC put together the representation within the same segment while distancing them from others, resulting in a representation that reflects the comprehension of patterns. The results show that our method outperforms existing approaches in self-supervised time series representation learning for trajectory classification and clustering tasks and resolves various utilization issues. This paper's contributions can be listed as follows: 

\begin{itemize}
    \item To our knowledge, we proposed, for the first time, a novel self-supervised contrastive time series representation learning framework for multivariate aerospace trajectory.
    \item We propose an effective concept of similarity within trajectory data using segmentation from the iterative Ramer-Douglas-Peucker (RDP) algorithm. We have integrated the capabilities of a causal language model to generate a representation where each timestep collects all preceding information, making it suitable for real-time monitoring and tracing incomplete and variable-length trajectories.
    \item We introduced labeled trajectory datasets essential for assessing classification accuracy and evaluating clustering results' fidelity to the labels, an often overlooked aspect in prior works.
    \item ATSCC has proven exceptionally suitable for aircraft trajectory data, outperforming state-of-the-art baselines. The model has overcome limitations associated with locational interpretation by allowing analysis with the semantic representation.
\end{itemize}

\section{RELATED WORKS}

\subsection{Time Series Representation Learning} 
Several studies have proposed models that generate time series representation vectors by leveraging the similarity among samples. Random Warping Series (RWS) \cite{RWS} demonstrated a kernel method based on Dynamic Time Warping (DTW), capable of generating a time series representation using random features approximation. SPIRAL \cite{SPIRAL} proposed a method that converts a set of time series into vectors by solving non-convex and nom-deterministic polynomial-time hard optimization, preserving the pairwise similarity of instances.

Autoencoders are the most widely applied representation learning for the aerospace domain. An encoder compresses the data sample into a representation vector, which is then used by the decoder for reconstruction. The model is trained with a reconstruction loss formulated using arbitrary differentiable distance functions and can be constructed with various architectures such as multi-layer perceptrons (MLP) \cite{ZengDAE2021, Chu2022EnsembleClustering}, convolutional neural networks (CNN) \cite{Olive2020DeepTrajectory, Liu2023}, recurrent neural networks (RNN) \cite{TimeNet}, long-short term memory (LSTM) \cite{Liu2022DeepTrajectory, Fan2023TALSTM}, dilated causal convolutional networks \cite{WaveNet}, as implemented by \cite{TCNAE_Mo, TCNAE_Thill}, or transformers \cite{Vaswani2017Transformer} as in \cite{tevet2022motionclip}.

Contrastive learning optimizes the encoded representation in the embedding space, ensuring the positive pairs or groups are similar in the embedding space while being distinct from the negative ones. The main idea of contrastive representation learning is to develop an effective definition of positive and negative samples, architecture, and training strategy. T-loss \cite{TLoss} employs a triplet loss for training, where a positive sample is obtained from a sub-series within the random reference area of an instance, and the negatives are all other sub-series. TNC \cite{TNC} leverages the local smoothness of the time series to model a positive temporal neighborhood, while the negatives are sub-series that are temporally distant. TNC maximizes similarity likelihood within the neighborhood while minimizing that of the negatives. TS-TCC \cite{TSTCC} maximizes the agreement between two augmented views of an instance while setting them apart from other samples. TS2Vec \cite{TS2Vec} employs random cropping and binomial masking for data augmentation. The encoder is trained by applying the hierarchical contrastive loss on the embedding of two cropped time series, both temporally and instance-wise, at every temporal max pooling level. InfoTS \cite{InfoTS} incorporates meta-learning techniques to identify the most suitable time series augmentation method. The encoder is trained using InfoNCE loss, contrasting temporally and instance-wise, while the meta-learner learns by balancing fidelity and variety loss, similar to \cite{Tian2020Infomin}. Contrastive learning allows us to define semantic similarities and rules for elements in data, aligning data representation more closely with contextual understanding.

\subsection{Trajectory Classification}
One of the fundamental challenges in trajectory recognition is runway classification \cite{Bosson2018}. The study demonstrated the classification of landing trajectories using a labeled dataset, comparing various classifiers. However, recent works on traffic patterns are relatively fewer than those on clustering, as public trajectory data are often unlabeled. Efforts for data labeling have been demonstrated in several works, where clustering results are analyzed and used as labels to train classifiers such as the Random Forest algorithm \cite{Murca2016, Murca2018, Madar2021Euclidean}, MLP \cite{Campfens2020Classification}, and LSTM \cite{Deng2022}. Directly enforcing a classification loss such as cross-entropy tends to result in representations that are close to one-hot vectors. Moreover, these works emphasize the necessity of labeled datasets.

\subsection{Trajectory Clustering}
In the simplest way, trajectories can be resampled to a fixed length and then clustered using Euclidean distance, as demonstrated in \cite{Murca2016, Murca2018, Madar2021Euclidean, Basora2017}, or using weighted Euclidean distance, as shown in \cite{Corrado2020WeightedEuclidean}. For variable-length datasets, clustering can be performed using a matrix of time series pairwise distances. Examples include DTW \cite{SakoeDTW1978} combined with agglomerative clustering \cite{Agglo} as in \cite{Deng2022, Rehm2010ClusteringOfFlightTracks}, route similarity \cite{Andrienko2013} with progressive clustering \cite{Rinzivillo2008} using OPTICS \cite{OPTICS} as demonstrated in \cite{Andrienko2018Relevant}, and Symmetrized Segment-Path Distance (SSPD) \cite{Besse2016SSPD} with HDBSCAN \cite{HDBSCAN} on simplified trajectories using the RDP algorithm \cite{Ramer1972, DOUGLAS1973} as shown in \cite{Basora2017}. Aircraft flight paths can also be a feature for clustering. For example, DTW combined with HDBSCAN \cite{HDBSCAN} was used on track angle sequences \cite{Gui2021Track}, while in \cite{Mcfadyen2016Circular}, track sequences were modeled using Von Mises distributions and clustered with K-medoids \cite{K-medoids} using Bhattacharyya distance \cite{Bhattacharyya}.

The partitioning-based method can be traced back to TRACLUS \cite{Lee2007TRACLUS}, which identifies points where states change rapidly for partitioning, uses DBSCAN \cite{DBSCAN} with line segment distance to cluster line segments, and then combines these segments to form clusters. Similarly, the methods in \cite{Gariel2011, Olive2019Clus} identify turning points based on heading changes and group those points using K-means \cite{Kmeans} and DBSCAN. These approaches then form the clusters by aggregating the sequences of waypoints from a dependency tree. These methods often rely on spatial interpretations and often result in a large number of clusters. The hidden Markov model in \cite{Chakrabarti2023Modeling} converts trajectories into string sequences of turning actions, which are then clustered using K-medoids with the edit distance.

Clustering on raw samples can be inefficient with high-dimensional samples. Many works have demonstrated clustering on the dimensionality-reduced samples, such as PCA \cite{Gariel2011, Eerland2016, Liu2023}, and t-SNE \cite{Olive2020DeepTrajectory, Damanik2022}; however, these methods are typically used for visualization. Moreover, they focus on preserving the dataset's characteristics rather than those of data instances. For representation-based methods, autoencoders are commonly used, and the trained encoder's output serves as the trajectory representation; for example, the MLP with Gaussian Mixture Model (GMM) \cite{GMM} in \cite{ZengDAE2021} or with K-Means in \cite{Chu2022EnsembleClustering}, and the CNN for visual representation with GMM as in \cite{Liu2023}. Deep clustering is typically an extension of the representation-based methods, often utilizing either a pretrained feature extractor or learning the representation concurrently with the training process. A popular framework for trajectory deep clustering is Deep Embedded Clustering (DEC) \cite{xie2016DEC}, implemented with various architecture choices such as CNN in \cite{Olive2020DeepTrajectory}, bidirectional LSTM in \cite{Liu2022DeepTrajectory}, and time and attribute LSTM (TA-LSTM), an LSTM with a time regulator gate for irregular time intervals, as in \cite{Fan2023TALSTM}. For both clustering on representations and deep clustering, having a high-fidelity representation can enhance performance by effectively converging into distinct classes.

\section{METHODOLOGY}

\begin{figure*}[htbp]
  \centering
  \includegraphics[width=0.95\textwidth]{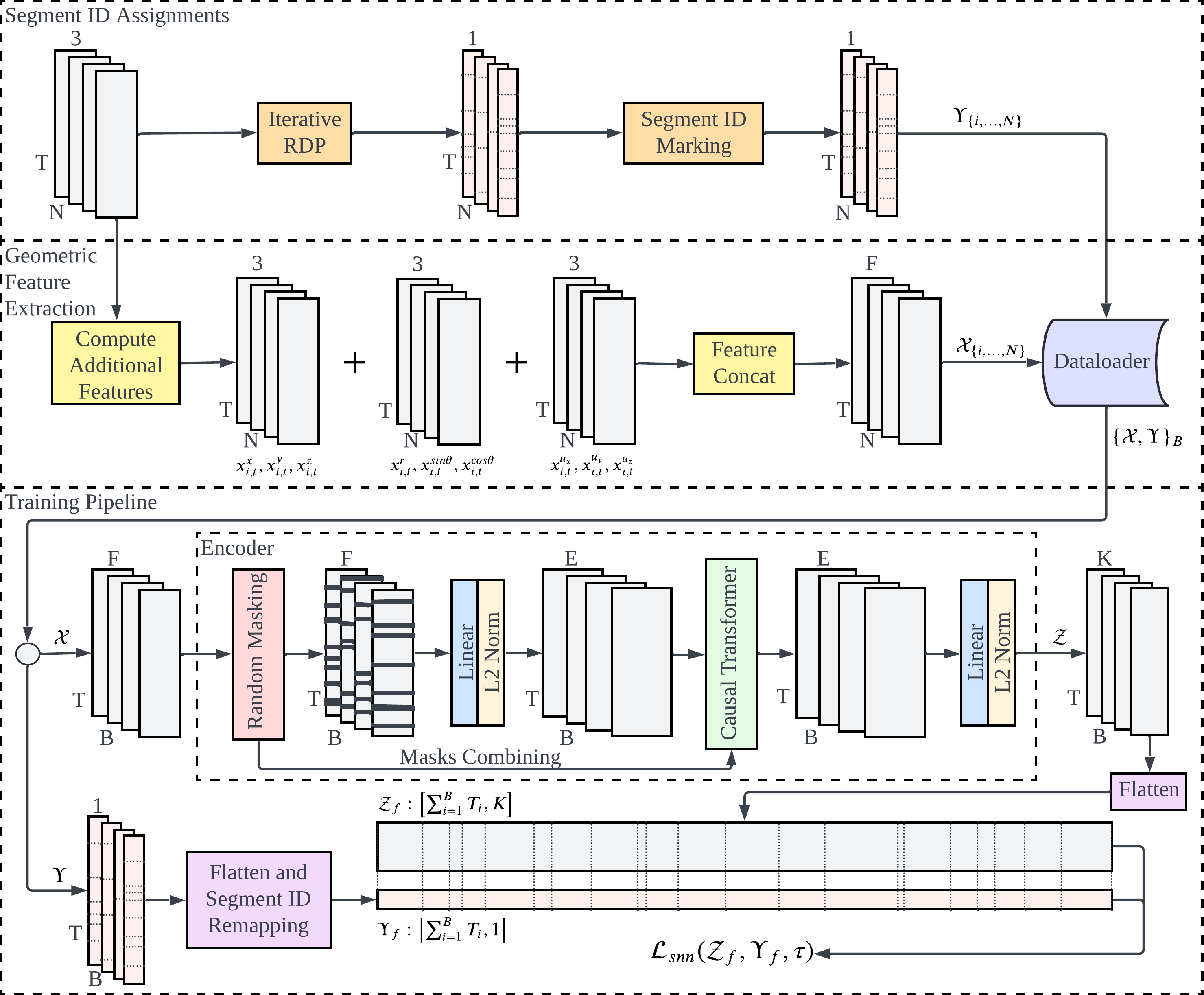}  
  \vspace{5pt}
  \caption{Diagram of the Proposed Framework for Aircraft Trajectory Segmentation-based Contrastive Coding (ATSCC): Includes Segment ID Assignments Using the Iterative RDP Algorithm, Geometric Feature Extraction, Training Pipeline with a Causal Transformer Encoder, ID Remapping, and Training with Soft Nearest Neighbor Loss.}
  \label{fig:ATSCC}
\end{figure*}

\subsection{Problem Definition}
For $N$ instances of time series in a variable length trajectory dataset, a multivariate time series, denoted as $\mathcal{X}_i = \{x_{i,1}, x_{i,2}, x_{i,3}, \dots, x_{i,T_i}\}$, has a dimension of $T_i \times F$, where $T_i$ is the sequence length of $i^{th}$ sample and $F$ is the number of features which varies based on the selected geometric features. In detail, \(x_{i,t}\) initially consists of position \(\{x_{i,t}^x, x_{i,t}^y, x_{i,t}^z\}\) in the East-North-Up (ENU) coordinates. In this paper, it can be extended up to 9 features, incorporating directional vectors \(\{x_{i,t}^{u_x}, x_{i,t}^{u_y}, x_{i,t}^{u_z}\}\) and polar components \(\{x_{i,t}^r, x_{i,t}^{\sin\theta}, x_{i,t}^{\cos\theta}\}\). It should be noted that the sequence length $T_i$ mentioned is the explicit time, and $T_i$ varies among instances as we deliberately avoided interpolating the trajectory data to a uniform length. The encoder $f_w (\mathcal{X}_{i})$, with learnable parameter $w$, is trained to best describe $\mathcal{X}_i$ with the representation $\mathcal{Z}_i = \{z_{i,1}, z_{i,2}, z_{i,3}, \dots, z_{i,T_i}\}$, that has a dimension of $T_i \times K$. Each $z_{i,t} \in \mathbb{R}^K$ simmarizes $\mathcal{X}'_{i,t} = \{x_{i,k} : k \leq t\}$ where $\mathcal{X}'_{i,t} \subseteq \mathcal{X}_i$. The training of the encoder $f_w (\mathcal{X}_{i})$ involves the assignment of the local segment ID denoted as $\Upsilon_i = \{\upsilon_{i,1}, \upsilon_{i,2}, \upsilon_{i,3}, \dots, \upsilon_{i,T_i}\}$ on each $\mathcal{X}_{i}$ as the notation for the loss function.

\subsection{Segment ID Assignments}
Many air traffic control (ATC) tasks exhibit segmenting behavior. For example, aircraft vectoring instructions remain unchanged until the aircraft reaches a designated turning position or is given a new instruction. In waypoint navigation, a sequence of waypoints instructs the aircraft operator to comply, forming the trajectory. In TBO, planned trajectories are divided by detailed checkpoints. Inspired by these tasks, we infer the semantic context within segments partitioned by significant positions.

The ATSCC framework (Fig \ref{fig:ATSCC}), inspired by these ATC tasks, utilizes the geometric properties of trajectories instead of data augmentation, which has been debated for its infeasibility across diverse datasets due to their unique temporal dynamics \cite{InfoTS}. Specifically, trajectories can be segmented into significant points \cite{Gariel2011, Olive2019Clus}, which are assumed to be close enough to the instructed waypoints, thus forming segments within the trajectory. Using this characteristic, the ATSCC framework aims to group together the representations at timestamps within a segment while distancing them from those at timestamps outside that segment and from those within other instances.

The RDP algorithm \cite{Ramer1972, DOUGLAS1973} reduces points in a curve while preserving its essential shape and geometric integrity, resulting in a simplified curve with timestamps of significant points that segment the curve. We used an open-source Python library for the iterative RDP algorithm \cite{RDPcode}, as described in Algorithm \ref{alg:RDP}. It starts by marking the beginning and end points of the curve as significant. Then, for each segment between significant points, the algorithm identifies points that exceed a perpendicular distance threshold, $\epsilon$, from the line segment and marks these points as significant. This process repeats until no further points exceed the threshold. The perpendicular distance from a point \( x_{i,k} \) to a line segment defined by points \( x_{i,s} \) and \( x_{i,e} \), is denoted as \( d(x_{i,k}, x_{i,s}, x_{i,e}) \) and can be calculated as follows:
\begin{equation}
d(x_{i,k}, x_{i,s}, x_{i,e}) = 
\begin{cases} 
\| x_{i,k} - x_{i,s} \| & \hspace{-15pt} \text{if } x_{i,s} = x_{i,e},\\
\frac{\left| \| (x_{i,e} - x_{i,s}) \times (x_{i,s} - x_{i,k}) \| \right|}{\| x_{i,e} - x_{i,s} \|} & \text{otherwise}.
\end{cases}
\end{equation}
Here, $s$, $k$, and $e$ denote the indices of the start, considering, and end points of the trajectory segment, respectively. \( x_{i,s} \), \( x_{i,k} \) and \( x_{i,e} \) refer to thier corresponding points within $\mathcal{X}_i$. Note that for this algorithm, $x_{i,t} = \{x_{i,t}^x, x_{i,t}^y, x_{i,t}^z\}$. The iterative RDP algorithm produces the mask \( M_i = \{ m_{i,1}, m_{i,2}, m_{i,3}, \dots, m_{i,T} \} \), where each \( m_{i,t} \in \{0, 1\} \). An instance's segment IDs $\Upsilon_i = \{ \upsilon_{i,1}, \upsilon_{i,2}, \upsilon_{i,3}, \ldots, \upsilon_{i,T} \}$ is obtained from the cumulative sum by each segment ID $\upsilon_{i,t} = \sum_{k=1}^t m_{i,k}$ and $\upsilon_{i, T} = \upsilon_{i, T-1}$ where $\upsilon_{i,t} \in \mathbb{Z}$ is marked corresponded to the temporal indices, $t$ on $x_{i,t}$.

The process of segment ID assignment can be described as a function where $\Upsilon_i = RDP(\mathcal{X}_i, \epsilon)$, indicating that $\Upsilon_i$ is the result of performing segment ID assignment via the RDP algorithm with an allowable perpendicular line segment distance of $\epsilon$ to the trajectory $\mathcal{X}_i$. $\epsilon$ serves as a measure of the allowable error between $\mathcal{X}_i$ and its simplified version. Thus, changing in $\epsilon$ changes how the operational context boundary is defined, so $\epsilon$ is taken as a hyperparameter of the ATSCC. All $\Upsilon_i$ for $N$ instances of $\mathcal{X}_i$ are precomputed for computational efficiency. They are padded with NaN to $T_{\text{max}} = \max_{i \in 1 \dots N}(T_i)$, in the same manner as the unequal length trajectory dataset, and stored within the PyTorch Dataloader module for later use in the training pipeline.

\begin{algorithm}
\caption{Local Label Marking via Iterative Ramer-Douglas-Peucker Algorithm}
\begin{algorithmic}[1]
\Procedure{RDP}{$\mathcal{X}_i$, $\epsilon$}
    \State // \textit{Step 1: Iterative RDP Algorithm \cite{RDPcode}}
    \State $\mathcal{S} \gets \{(1, T_i)\}$  
    \State $M_i \gets \{1 : k = 1, \dots, T_i\}$  
    \State $n_{\mathcal{S}} = |\mathcal{S}|$
    \While{$n_{\mathcal{S}} \not = 0$} 
        \State $(s, e) \gets \mathcal{S}[n_{\mathcal{S}}]$ 
        \State $\mathcal{S} \gets \mathcal{S} - \{(s,e)\}$
        \State $d_{max} \gets 0.0$  
        \State $t \gets s$  

        \For{$k \gets t + 1$ \text{ to } $t_{e} - 1$}
            \If{$m_{i,k}$}  
                \State $d \gets d(x_{i,k}, x_{i,s}, x_{i,e})$  
                \If{$d > d_{max}$}  
                    \State $t \gets k$  
                    \State $d_{max} \gets d$  
                \EndIf
            \EndIf
        \EndFor

        \If{$d_{max} > \epsilon$}  
            \State $\mathcal{S} \gets \mathcal{S} + \{(s,t),(t,e)\}$ 
        \Else
            \For{$k \gets s + 1$ \text{ to } $e - 1$}
                \State $m_{i,k} \gets 0$  
            \EndFor
        \EndIf
        \State $n_{\mathcal{S}} = |\mathcal{S}|$
    \EndWhile

    \State // \textit{Step 2: Local segment ID assignment}
    \State $\Upsilon_i \gets \{\}$
    \For{$t$ in $T$}
        \If{$t \not = T$}
            \State $\upsilon_{i,t} = \sum_{k=1}^t m_{i,k}$
        \Else
            \State $\upsilon_{i,t} = \sum_{k=1}^{t-1} m_{i,k}$
        \EndIf
        \State $\Upsilon_i \gets \Upsilon_i + \{\upsilon_{i,t}\}$
    \EndFor
    
    \State \Return $\Upsilon_i$  
\EndProcedure
\end{algorithmic}
\label{alg:RDP}
\end{algorithm}

\subsection{Geometric Feature Extraction}
The trajectory data describe the sequence of aircraft's positions in ENU coordinates as $\mathcal{X}_i \in \{ x_{i,1}, x_{i,2}, x_{i,3}, \dots, x_{i,T}\}$; each $x_{i,t}$ initially consists of $x_{i,t}^x, x_{i,t}^y, x_{i,t}^z$. To extend the underlying information in trajectory states, this feature extraction calculates additional geometric features prior to the training stage. Since contrastive representation learning can handle features of different scales, these features enhance information without typical scaling issues found in feature-based methods or autoencoders. Empirically, this inclusion improves performance, as discussed in the ablation study.

Aircraft path angle is a useful feature for representing trajectory shape \cite{Chu2022EnsembleClustering, Gui2021Track, Mcfadyen2016Circular}. Moreover, most ATM instructions involve path angles, such as headings, glide paths, or directing aircraft between waypoints and specific altitudes. In other words, ATC uses not only positions but also the direction of the aircraft. This paper uses the directional unit vector to represent aircraft paths, avoiding trigonometric singularities of angles. The aircraft's path vector at each timestamp is given by

\begin{equation}
    \{x_{i,t}^{u_x}, x_{i,t}^{u_y}, x_{i,t}^{u_z}\} = \frac{x_{i, t+1} - x_{i,t}}{\left\| x_{i, t+1} - x_{i,t} \right\|},
\end{equation}
where \( \left\| x_{i, t+1} - x_{i,t} \right\| \) denotes the Euclidean norm of the vector difference, and \(x_{i,t} = \{x_{i,t}^x, x_{i,t}^y, x_{i,t}^z\}\). For the final time step \( T_i \), it is assumed that without further updates, the aircraft maintains the last computed path, such that 
\begin{equation}
    \{x_{i,T_i}^{u_x}, x_{i,T_i}^{u_y}, x_{i,T_i}^{u_z}\} = \{x_{i,T_i-1}^{u_x}, x_{i,T_i-1}^{u_y}, x_{i,T_i-1}^{u_z}\}.
\end{equation}

Interpreting positions in polar coordinates centered at reference waypoints is crucial for navigational aids such as VOR/DME (Very High-Frequency Omnidirectional Range with Distance Measuring Equipment) stations. The sensitivity of these states increases as the aircraft maneuvers near the reference point, in this case, the airport. This behavior establishes rapid changes in features, providing ease of recognition. The polar components feature at each timestamp can be calculated with,
\begin{equation}
    \{x_{i,t}^{r}, x_{i,t}^{\sin\theta}, x_{i,t}^{\cos\theta}\} = \left\{\begin{matrix}
    \sqrt{\left(x_{i,t}^x\right)^2 + \left(x_{i,t}^y\right)^2} \\ 
    \sin\left(\arctan2(x_{i,t}^y, x_{i,t}^x)\right) \\ 
    \cos\left(\arctan2(x_{i,t}^y, x_{i,t}^x)\right)
    \end{matrix}\right\}^T.
\end{equation}

We use the sine and cosine of the angle from the $\arctan2$ function to avoid singularities and discontinuities at 0 and 360 degrees. This method ensures a smooth transition across quadrant boundaries. The features are combined to form \(x_{i,t} = \{ x_{i,t}^x, x_{i,t}^y, x_{i,t}^z, x_{i,t}^{u_x}, x_{i,t}^{u_y}, x_{i,t}^{u_z}, x_{i,t}^r, x_{i,t}^{\sin\theta}, x_{i,t}^{\cos\theta} \}\). These are then stored in the PyTorch DataLoader module along with the segment IDs \(\Upsilon_i\).

\subsection{Encoder Architecture}
The encoder \(f_w\) consists of a random masking module, a Causal Transformer backbone, and input/output projection layers, as shown in Fig \ref{fig:ATSCC}. The encoder first applies binomial timestamp masking to the inputs during training, similar to \cite{TS2Vec}, but our encoder employs masking before the input projection to mimic missing states caused by irregular transmissions, providing regularization like dropout. This random mask is later merged with the source padding mask and causal mask for attention to ignore the masked tokens.

\begin{figure}[htbp]
  \centering
  \includegraphics[width=0.85\linewidth]{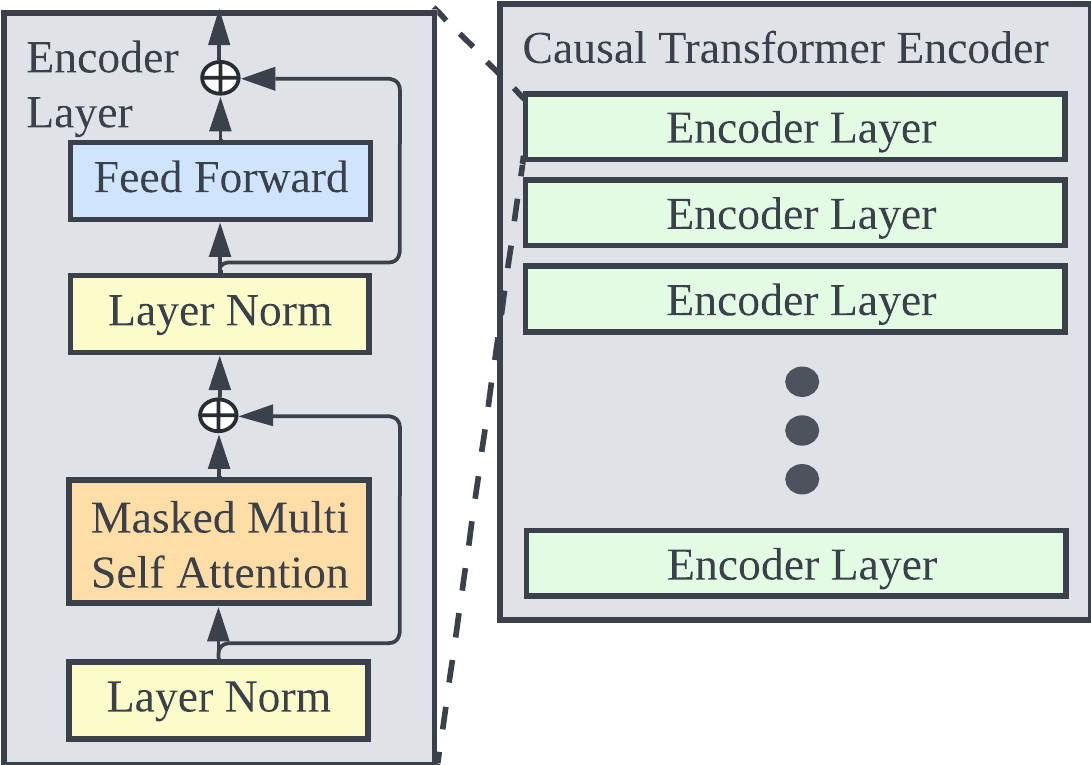}
  \vspace{5pt}
  \caption{Architecture of Causal Transformer Encoder}
  \label{fig:encoder}
\end{figure}

The input linear projection expands each state \(x_{i,t}\) from dimension \(K\) to the token dimension \(E\) of the encoder backbone. The architecture of the Causal Transformer backbone, illustrated in Fig \ref{fig:encoder}, is similar to the designs of GPT-2 \cite{GPT2} and GPT-3 \cite{GPT3}. For each batch of inputs $\mathcal{X} = \{\mathcal{X}_1, \mathcal{X}_2, \mathcal{X}_3, \dots, \mathcal{X}_B\}$, the encoder handle unequal-length sequences by applying key padding masks on inputs, outputs, and attention matrices. The causal attention mechanism restricts attention only to previous tokens, ensuring that the output for any position reflects the set encoding of all preceding tokens. Thus, the model's output at any given time, \(z_{i,t}\), is the summary of \(\mathcal{X}'_{i,t} = \{x_{i,k}:k\leq t\}\). Each encoder state, having a dimension of \(E\), is then linearly projected into \(z_{i,t}\) with a representation size of \(K\).

We follow the NoPos approach \cite{NoPos}, which shows that the transformer causal language model (CLM) can perceive positional information without explicit positional encoding. Thus, when applied to time series data, the CLM effectively captures their sequential behavior. We apply L2 normalization after each linear projection layer as suggested in \cite{Hemisphere} that it can improve performance by making the embeddings linearly separable. Additionally, L2 normalization on tokens before encoding slightly enhances the learned representation's effectiveness for downstream tasks.

\subsection{Training Pipeline}
As the ATSCC framework defines contextual similarity using the segment IDs assigned by the RDP algorithm, this contrastive representation learning becomes simple yet effective. First, a batch time series data, $\mathcal{X} = \{ \mathcal{X}_1, \mathcal{X}_2, \mathcal{X}_3, \dots, \mathcal{X}_B \}$, is encoded through the encoder $f_w$, resulting in a batch of representation $\mathcal{Z} = \{ \mathcal{Z}_1, \mathcal{Z}_2, \mathcal{Z}_3, \dots, \mathcal{Z}_B \}$. As $z_{i,t}$ represents $\mathcal{X}'_{i,t} = \{x_{i,k} : k \leq t\}$, the contrasting applied to each element of an encoded batch \(\mathcal{Z}\) is equivalent to both temporal and instance-wise contrasting of a set the incomplete instances $\{\mathcal{X}'_{i,t}\}$ for \(t \in \{1, 2, 3, \dots, T_i\}\) and \(i \in \{1, 2, 3, \dots, B\}\). The contrasting on an encoded batch \(\mathcal{Z}\) is dictated by the coresponding batch of segment IDs, \(\Upsilon = \{ \Upsilon_1, \Upsilon_2, \Upsilon_3, \dots, \Upsilon_B \}\). The framework define positive samples as \(z_{i,t}\) within the same segment that shares the same \(\upsilon_{i,t}\), while treating other \(z_{i,t}\) in different segments and \(z_{j,t}\) from other instances \(j \in \{1, 2, 3, \dots, B\}\) with \(j \neq i\) as negative samples. The procedure of loss calculation is described in Algorithm \ref{alg:lossfunc}.

\begin{algorithm}
\caption{Calculation of Soft Nearest Neighbor Loss}
\label{alg:lossfunc}
\begin{algorithmic}[1]
\Procedure{SNNL}{$\mathcal{Z}, \Upsilon, \tau$}
    \State $\mathcal{Z}_{f} \gets \{\}$ 
    \State $\Upsilon_{f} \gets \{\}$
    \State $\upsilon_{\text{current}} \gets 1$
    \For{$i \gets 1 \text{ to } B$}  
        \State // \textit{Flatten $\mathcal{Z}$}
        \State $K \gets \{t | \upsilon_{i,t} \not = \text{NaN}, \upsilon_{i,t} \in \Upsilon_i\}$
        \State $\mathcal{Z}_{f} \gets \mathcal{Z}_{f} + \{z_{i,t}|t \in K, z_{i,t} \in \mathcal{Z}_i\}$
        \State // \textit{Segment ID remapping} 
        \State $g_{m} : \{\} \rightarrow \{\}$
        \State $\Upsilon_{i,\text{new}} \gets \{\}$
        \State $\Upsilon_i \gets \{\upsilon_{i,t}|t \in K, \upsilon_{i,t} \in \Upsilon_i\}$  
        \For{$\upsilon_{i,t}$ \textbf{in} $\Upsilon_i$}  
            \If{$g_m(\upsilon_{i,t}) \text{ does not exist}$}
                \State $g_m(\upsilon_{i,t}) \gets \upsilon_{\text{current}}$
                \State $\upsilon_{\text{current}} \gets \upsilon_{\text{current}} + 1$
            \EndIf
            \State $\Upsilon_{i,\text{new}} \gets \Upsilon_{i,\text{new}} + \{g_m(\upsilon_{i,t})\}$
        \EndFor
        \State $\Upsilon_{f} \gets \Upsilon_{f} + \Upsilon_{i,\text{new}}$
    \EndFor
    \State // \textit{Calculate soft-nearest neighbor loss}
    \State $\text{Loss} = \mathcal{L}_{ssn}(\mathcal{Z}_{f}, \Upsilon_{f}, \tau)$; Equation (\ref{eq:snnlnew})
    \State \Return Loss
\EndProcedure
\end{algorithmic}
\end{algorithm}

The batch of embedding $\mathcal{Z}$ is flattened into 2-dimensional $\mathcal{Z}_f$ having the number of features of $F$ and the length of $\sum_{i=1}^B T_i$, for $T_i$ is not equal across instances. During the ID assigning, each sequence of segment IDs $\Upsilon_i$ is marked locally, starting from 1. The ID remapping on the batch of segment IDs $\Upsilon$ is performed using the updatable mapping function $g_m$. This procedure ensures that when local segment IDs are flattened, the uniqueness of IDs across segments is guaranteed, preventing segments from different samples from being mistakenly identified as positives. The length of the flattened segment IDs, $\Upsilon_i$ is $\sum_{i=1}^B T_i$, where each ID $\upsilon_{i,t}$ matches the corresponding representation vector $z_{i,t}$. Given the presence of multiple positive samples, the ATSCC framework employs a modified soft-nearest neighbor loss \cite{frosst2019analyzingSNNL}, which utilizes scaled vector multiplication instead of distances, as each $z_{i,t}$ has been L2 normalized. The rearranged soft-nearest neighbor loss is given by equation (\ref{eq:snnlrearrange}).

\begin{multline}
    \mathcal{L}_{snn}(\mathcal{Z}_{f}, \Upsilon_{f}, \tau) = \\
    -\mathop{{}\mathbb{E}}_{\substack{z \sim \mathcal{Z}_{f} \\ \upsilon \sim \Upsilon_{f} \\ i \sim |\mathcal{Z}_f|}} \left( \log\sum_{\substack{j = 1 \\ j \neq i \\ \upsilon_i = \upsilon_j}}^{|\mathcal{Z}_f|} e^{\left(\frac{z_i^T z_j}{\tau}\right)} - \log\sum_{\substack{k = 1 \\ k \neq i}}^{|\mathcal{Z}_f|} e^{\left(\frac{z_i^T z_k}{\tau}\right)}\right).
\label{eq:snnlrearrange}
\end{multline}
\vspace{4pt}

A single model parameter update step in ATSCC processes a large number of $z_{i,t}$ in one batch; for example, a batch size of 16 trajectories with an average length of 500 steps results in 8,000 representation vectors in $\mathcal{Z}_{f}$. We assume that the inclusion of positive contrasting terms in the sum of negative terms has a minimal effect. Moreover, omitting positive contrasts in the negative terms, as in equation (\ref{eq:snnlnew}), leads to empirically better performance.

\begin{multline}
    \mathcal{L}_{snn}(\mathcal{Z}_{f}, \Upsilon_{f}, \tau) = \\
    -\mathop{{}\mathbb{E}}_{\substack{z \sim \mathcal{Z}_{f} \\ \upsilon \sim \Upsilon_{f} \\ i \sim |\mathcal{Z}_f|}} \left( \log\sum_{\substack{j = 1 \\ j \neq i \\ \upsilon_i = \upsilon_j}}^{|\mathcal{Z}_f|} e^{\left(\frac{z_i^T z_j}{\tau}\right)} - \log\sum_{\substack{k = 1 \\ k \neq i \\ \upsilon_i \neq \upsilon_k}}^{|\mathcal{Z}_f|} e^{\left(\frac{z_i^T z_k}{\tau}\right)}\right)
\label{eq:snnlnew}
\end{multline}
\vspace{4pt}

The loss function maximizes agreement between the representation vectors in the assigned segments while explicitly differentiating them from those in other segments and other instances in a batch. This modified form ensures that enforcing the soft-nearest neighbor loss supports our fundamental motivation; that is, the collective trajectory states within the same operational context area should result in a similar representation clearly distinct from those outside.

\begin{figure*}[htbp]
  \centering
  \includegraphics[width=0.825\textwidth]{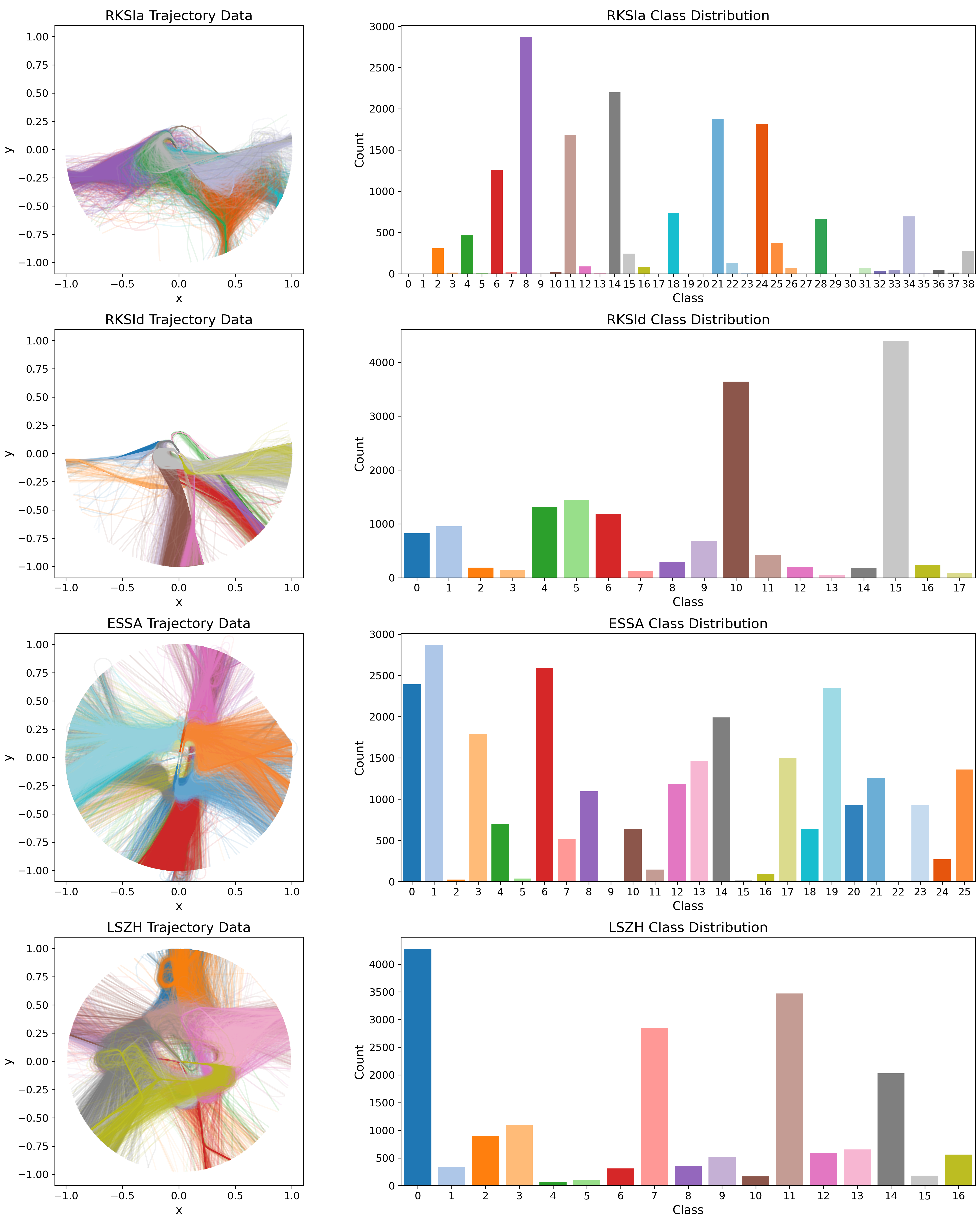}  
  \caption{Visualization of Trajectory Datasets and Histogram of Class Distribution (by order): Incheon Airport Arrival and Departure Trajectories, Stockholm Arlanda Airport Arrival Trajectories, and Zurich Airport Arrival Trajectories, with Corresponding Histograms of Class Distribution.}
  \label{fig:classdist}
\end{figure*}

\section{RESULTS AND DISCUSSION}
In this section, we first elaborate on the dataset preparation used for training and evaluation. We outline the experiment setting for evaluating the learned trajectory representation on fundamental air traffic management downstream tasks, including classification and clustering. We compare the performance of our learned representation with state-of-the-art representation learning methods and autoencoders to demonstrate ATSCC's suitability for air traffic data. The feasibility of model components was analyzed in the ablation study. Additionally, we provide a detailed visual explanation of the learned representation. We further demonstrate the utilization of ATSCC for trajectory categorization, addressing a common problem in the field.

\subsection{Datasets}
\subsubsection{Dataset Sources}
This study used trajectory data from three different airports, totaling four datasets. We aim to demonstrate the broad applicability of the ATSCC framework across various airport configurations. This paper uses data from the authors and other researchers to ensure the framework's applicability, reliability of datasets, and experiment validity. The data sources are:
\begin{itemize}
    \item \textbf{Incheon International Airport} (Denoted as ICAO code: RKSI): The ADS-B data were sourced from the Opensky database \cite{Schafer2014OpenSky}. The data for flights from 2018 to 2023 were queried using the flight identification numbers from the Airportal website, \cite{MLIT2024Airportal}. We downsampled the trajectory data from the southbound and southeastbound flights to improve balance. The datasets for arrivals and departures are denoted as RKSIa and RKSId, respectively.
    \item \textbf{Stockholm Arlanda Airport} (Denoted as ICAO code: ESSA): The data is from the Swedish Civil Air Traffic Control (SCAT) dataset \cite{Nilsson2023SCAT}, which includes surveillance data, weather, flight plans, and airspace data. We focused on surveillance data for positional states and flight plans to assist our manual labeling. We only use arrival data for this airport, as the departure data was found incomplete.
    \item \textbf{Zurich Airport} (Denoted as ICAO code: LSZH): The data was sourced from the \cite{Olive2020Dataset}, where the focus was on analyzing Zurich Airport flight arrivals.
\end{itemize}

\subsubsection{Dataset Preprocessing}
The dataset preprocessing involves data cleaning, transformation, resampling, smoothing, and scaling. We first omitted the preprocessing for incomplete trajectories. Then, we extracted positional states from the surveillance data, including latitude, longitude, and barometric altitude. Barometric altitude was chosen for its lower noise levels and greater reliability compared to GPS vertical accuracy \cite{GPSacc} and because it is commonly used for traffic control. The positional features were transformed into Cartesian vectors on the ENU (East-North-Up) coordinate centered at the airport. Then, trajectories were bounded by a radius, \( r_{\text{max}} \), determined by the farthest waypoints in the area chart of the Aeronautical Information Publication (AIP) with 120 km for Incheon airport, 100 km for Stockholm Arlanda airport, and 40 nautical miles for Zurich airport \cite{Olive2020Dataset}. The trajectories were resampled to 1-second intervals without interpolating into a fixed number of timestamps. Then, the outlier states were removed, and the Savizky-Golay filter \cite{Savitzky1964} was applied to smooth the trajectories. Each trajectory was scaled to the range \([-1, 1]\) by dividing by \(r_{max}\), such that \(\mathcal{X}_i \gets \mathcal{X}_{i}/r_{max}\). A preprocessed instance $\mathcal{X}_{i}$ has the states $x_{i,t} = \{x_{i,t}^x, x_{i,t}^y, x_{i,t}^z\}$. The trajectories are padded with NaN to $T_{\text{max}} = \max_{i \in 1 \dots N}(T_i)$. The training and testing sets were split with a 50\% proportion, as is common with many datasets; moreover, a significant amount of test data is needed to evaluate clusterability.


\subsubsection{Dataset Labeling}
The lack of labeled datasets restricts the advancement in developing classification models, which require training with true labels as done by \cite{Bosson2018}. The studies cited in \cite{Murca2016, Murca2018, Madar2021Euclidean, Deng2022, Campfens2020Classification} utilized clustering to generate labels; however, this method is commonly practiced in the artificial intelligence field as pseudo-labeling. Moreover, many clustering studies omit the labels in their evaluations, focusing on the proximity of instances within clusters \cite{ZengDAE2021, Chu2022EnsembleClustering, Liu2022DeepTrajectory, Damanik2022, Fan2023TALSTM}; nevertheless, the primary concern remains the fidelity of class assignments assessed using the true label. For these reasons, it is essential to have data labeled by humans, as in \cite{ImageNet, cocodataset}, to align with humans' semantic understanding of data.

The datasets were manually labeled using information published in the AIPs. Labeling arrival trajectories at Incheon Airport and Zurich Airport followed the same procedures. For arrival corridors labeling, performing K-means on the set of all $\{x^x_{i,1}, x^y_{i,1}\}$ for $i \in 1 \dots N$ provided a good initialization of segments before manual adjustments. The runways were labeled by manually drawing linear classification lines on $\{x^x_{i,T_i}, x^y_{i,T_i}\}$ for $ i \in 1 \dots N$, to classify the touchdown points. A similar manual classification on initial approach fixes (IAFs) was performed to classify the approach procedures. As a result, the classes are defined as the combination of runways, IAFs, and Standard Terminal Arrival Routes (STARs). For the Incheon departure dataset, labeling departure corridors mirrored that of arrivals but was implemented using the set $\{x^x_{i,T_i}, x^y_{i,T_i}\}$. In the AIP, the pairs of dependent runways comply with the same Standard Instrument Departure (SID); we labeled each $\{x^x_{i,1}, x^y_{i,1}\}$ as dependent runway pairs, using the same manual classification procedure. The labels for departure data denote the SIDs. The labeling of the trajectory data at Stockholm Arlanda airport was straightforward as the surveillance data were given with the flight plan data \cite{Nilsson2023SCAT}. The STARs and the situated runways are recorded; therefore, after minimal corrections and labeling with the initial approach fixes, the classes are given by combinations of runways, IAFs, and STARs.

Each set of class descriptions was then converted to integers denoted as $Y = \{ y_1, y_2, y_3, \dots y_N \}$ for $y_i \in \mathbb{Z}$. It is important to emphasize that these instance-level labels will not be used in representation learning via the self-supervised ATSCC framework or any baseline methods discussed in this paper; they are only for evaluation purposes. The trajectory datasets visualization with class distribution is illustrated in Fig \ref{fig:classdist}.


\subsection{Experiments}
\subsubsection{Hyperparameters and Reproduction Remarks}
Our model employs the CLM configuration from \cite{NoPos}, featuring 12 layers with a model dimension of 768, feed-forward dimensions of 3072 using Gaussian Error Linear Unit (GELU) activation, 12 attention heads with a 0.35 drop rate, and a binomial random masking probability set at 0.2. Following \cite{TLoss, TS2Vec}, the dimension of the representation $z_{i,t}$ is set at 320. The batch size, \( B \), is set at 16 for training with the AdamW optimizer, using a learning rate of \( 1 \times 10^{-5} \) and a weight decay \( \lambda \) for regularization set at \( 1 \times 10^{-5} \). For unscaled ATM trajectory data, we recommend using the RDP threshold as $\epsilon \cdot r_{max}$ for the maximum control radius $r_{max}$. However, as our data has been scaled to $[ -1, 1]$, we omitted the notation of $r_{max}$. We performed a grid search based on performance evaluation over the RDP threshold, $\epsilon \in \{0.0001, 0.001, 0.01, 0.1\}$ and the temperature in loss function $\tau \in \{0.01, 0.05, 0.1, 0.5, 1.0, 5.0, 10.0\}$. For ATSCC, the learned representation $z_{i,t}$ represents all its previous states; thus, the last timestamp's representation $z_{i,T_i}$ is extracted as the instance-level representation for evaluation. To reduce the computational burden, the trajectory dataset was downsampled every 5 seconds without standardization. To ensure the consistency of our results, we trained and evaluated the model with 5 different seeds. For all training and evaluation, this paper used Python 3.11.8 and PyTorch 2.0.1 with CUDA Toolkit 11.8, running on an Nvidia GeForce RTX 4090 GPU.

\subsubsection{Reproduction of Baselines}
This paper discusses the ATM trajectory recognition tasks at the instance level, including classification and clustering. In the same manner, as the training and evaluation of ATSCC, we downsampled the trajectories every 5 seconds and used identical geometric features across all baselines with $x_{i,t} = \{ x_{i,t}^x, x_{i,t}^y, x_{i,t}^z, x_{i,t}^{u_x}, x_{i,t}^{u_y}, x_{i,t}^{u_z}, x_{i,t}^r, x_{i,t}^{\sin\theta}, x_{i,t}^{\cos\theta} \}$. Following \cite{TLoss, TS2Vec}, the dimension of the representation vector is set to 320. We referred to the original code provided by the authors of the baseline papers to reproduce the results. The reproduction details are described as follows:

\begin{itemize}
    \item \textbf{SPIRAL} \cite{SPIRAL}: As an example of a non-neural network optimization-based method, we have reproduced SPIRAL using the DTW. We kept all parameters as default except for the representation size.
    \item \textbf{TCN-AE}: We use the Temporal Convolutional Network architecture from \cite{TLoss} to construct this autoencoder. Following \cite{TCNAE_Thill, TCNAE_Mo}, we employ the TCN as both the encoder and decoder with maxpooling and upsampling in the middle.
    \item \textbf{TF-AE}: We built a transformer autoencoder following the architecture in \cite{Vaswani2017Transformer}. The encoding involves feature extraction via a class token, while the decoding uses positional encoding as a time query on the latent vector for reconstruction, similar to \cite{tevet2022motionclip}.
    \item \textbf{T-Loss} \cite{TLoss} encourages the consistency of random subsequences using triplet loss. We used the default hyperparameters for the UCR and UEA datasets.
    \item \textbf{TNC} \cite{TNC} leverages local smoothness of time series and uses a sliding window to encode the data into shorter sequences, which we use to represent the instances, as aggregation to a single vector was not demonstrated. We used the waveform CNN encoder and HAR configuration. The size of the sub-series representation was set to 64, with a window size of $\text{int}(T_{max} / 20)$ and a sliding gap of 5.
    \item \textbf{TS-TCC} \cite{TSTCC} maximize the agreement between two augmented samples. We referred to the hyperparameters for the HAR Dataset. Following \cite{TS2Vec}, we exhibit the instance-level representation using Maxpooling on the base convolutional network.
    \item \textbf{TS2Vec} \cite{TS2Vec} maximize contextual consistency of time series using the hierarchical contrastive loss. We used the reproduction details the author demonstrated on the UCR and UEA datasets.
    \item \textbf{InfoTS} \cite{InfoTS} implements meta-learning technique in finding the best augmentation for representation learning. We reproduced it with the same hyperparameters used for the UCR and UEA datasets.
\end{itemize}

\begin{table*}
\caption{Performance of Self-Supervised Representation Learning Baselines: Evaluation of Learned Representations Across All Trajectory Datasets, Measured by Accuracy, NMI, and ARI Scores}
\centering
\begin{tblr}{
  cell{1}{2} = {c=3}{c},
  cell{1}{5} = {c=3}{c},
  cell{1}{8} = {c=3}{c},
  cell{1}{11} = {c=3}{c},
  cell{2}{2-13} = {c=1}{c},
  hline{1-3,5,8,15} = {-}{},
}
                                                                        & RKSIa                  &                 &                 & RKSId                  &                 &                 & ESSA            &                 &                 & LSZH            &                 &        \\
Baselines                                                               & ACC                    & NMI             & ARI             & ACC                    & NMI             & ARI             & ACC             & NMI             & ARI             & ACC             & NMI             & ARI    \\
Non-neural network                                                      &                        &                 &                 &                        &                 &                 &                 &                 &                 &                 &                 &        \\
\labelitemi\hspace{\dimexpr\labelsep+0.5\tabcolsep}SPIRAL \cite{SPIRAL} & 0.8344                 & 0.6211          & 0.3102          & 0.9946                 & 0.8732          & 0.7191          & 0.8503          & 0.7406          & 0.4352          & 0.9173          & 0.7920          & 0.6234 \\
Autoencoders                                                            &                        &                 &                 &                        &                 &                 &                 &                 &                 &                 &                 &        \\
\labelitemi\hspace{\dimexpr\labelsep+0.5\tabcolsep}TCN-AE               & 0.9753                 & 0.6682          & 0.3593          & 0.9982                 & 0.8706          & 0.6514          & 0.9942          & 0.7527          & 0.4420          & 0.9915          & 0.8060          & 0.6566 \\
\labelitemi\hspace{\dimexpr\labelsep+0.5\tabcolsep}TF-AE                & 0.9181                 & 0.6420          & 0.2881          & 0.9971                 & 0.8545          & 0.6229          & 0.9563          & 0.7489          & 0.4394          & 0.9674          & 0.7994          & 0.6159 \\
Contrastive Learning                                                    &                        &                 &                 &                        &                 &                 &                 &                 &                 &                 &                 &        \\
\labelitemi\hspace{\dimexpr\labelsep+0.5\tabcolsep}T-Loss \cite{TLoss}  & 0.9735                 & 0.6660          & 0.3238          & 0.9984                 & 0.8484          & 0.5969          & 0.9967          & 0.7552          & 0.4553          & 0.9916          & 0.8301          & 0.6349 \\
\labelitemi\hspace{\dimexpr\labelsep+0.5\tabcolsep}TNC \cite{TNC}       & 0.9728                 & 0.6982          & 0.4222          & 0.9980                 & 0.8784          & 0.6771          & 0.9904          & 0.7581          & 0.4837          & 0.9897          & 0.8983          & 0.7908 \\
\labelitemi\hspace{\dimexpr\labelsep+0.5\tabcolsep}TS-TCC \cite{TSTCC}  & 0.9847                 & 0.7207          & 0.4304          & 0.9980                 & 0.8831          & 0.6834          & 0.9989          & 0.7656          & 0.4670          & 0.9950          & 0.8667          & 0.7248 \\
\labelitemi\hspace{\dimexpr\labelsep+0.5\tabcolsep}TS2Vec \cite{TS2Vec} & 0.9786                 & 0.6890          & 0.3412          & 0.9984                 & 0.8660          & 0.6691          & 0.9981          & 0.7572          & 0.4463          & 0.9933          & 0.8373          & 0.6381 \\
\labelitemi\hspace{\dimexpr\labelsep+0.5\tabcolsep}InfoTS \cite{InfoTS} & 0.9764                 & 0.6668          & 0.3703          & 0.9986                 & 0.8758          & 0.6621          & 0.9986          & 0.7623          & 0.4608          & 0.9930          & 0.8604          & 0.6851 \\
\labelitemi\hspace{\dimexpr\labelsep+0.5\tabcolsep}ATSCC                & \textbf{0.9946}        & \textbf{0.8723} & \textbf{0.8195} & \textbf{0.9987}        & \textbf{0.9129} & \textbf{0.7723} & \textbf{0.9990} & \textbf{0.8640} & \textbf{0.6517} & \textbf{0.9977} & \textbf{0.9480} & \textbf{0.9037} 
\end{tblr}
\label{tab:comparison}
\end{table*}

\begin{figure*}[htbp] 
    \centering
    \begin{subfigure}{0.18\textwidth}
        \centering
        \includegraphics[width=\linewidth]{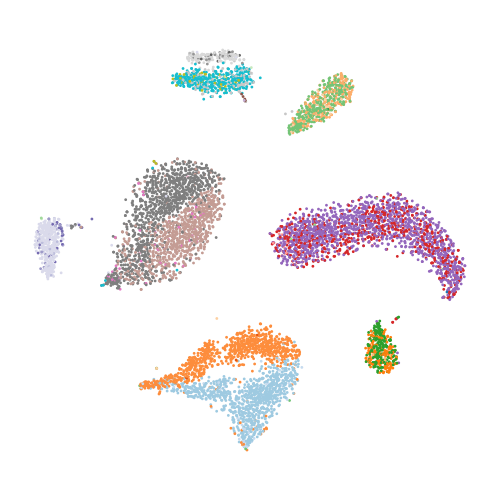}
        \caption{SPIRAL}
        \label{fig:RKSIaSPIRAL}
    \end{subfigure}\hfill
    \begin{subfigure}{0.18\textwidth}
        \centering
        \includegraphics[width=\linewidth]{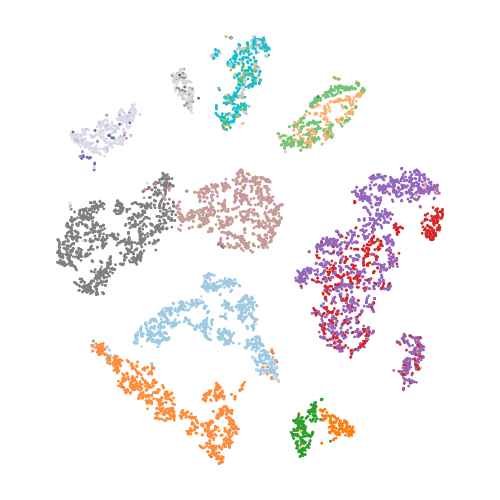}
        \caption{TCN-AE}
        \label{fig:RKSIaTCN-AE}
    \end{subfigure}\hfill
    \begin{subfigure}{0.18\textwidth}
        \centering
        \includegraphics[width=\linewidth]{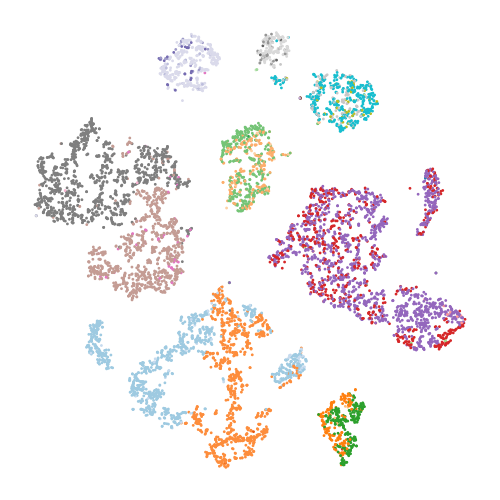}
        \caption{TF-AE}
        \label{fig:RKSIaTF-AE}
    \end{subfigure}\hfill
    \begin{subfigure}{0.18\textwidth}
        \centering
        \includegraphics[width=\linewidth]{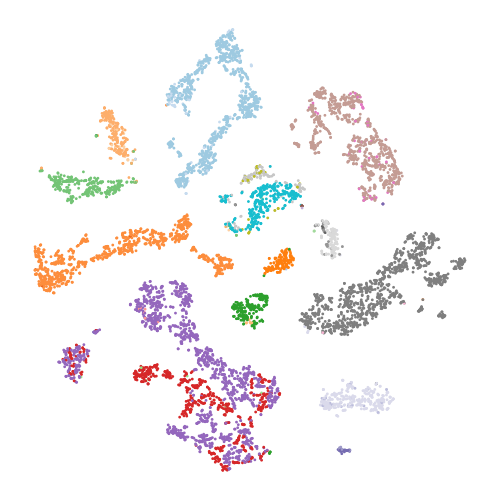}
        \caption{T-Loss}
        \label{fig:RKSIaT-Loss}
    \end{subfigure}\hfill
    \begin{subfigure}{0.18\textwidth}
        \centering
        \includegraphics[width=\linewidth]{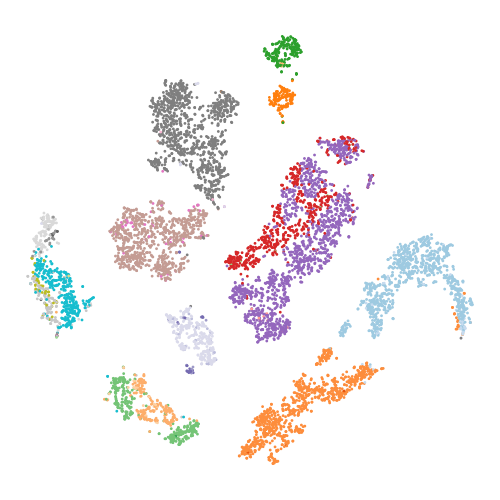}
        \caption{TNC}
        \label{fig:RKSIaTNC}
    \end{subfigure}

    \hspace{0.09\textwidth}
    \begin{subfigure}{0.18\textwidth}
        \centering
        \includegraphics[width=\linewidth]{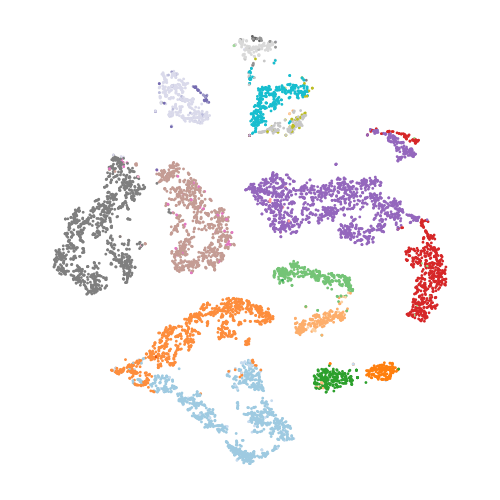}
        \caption{TS-TCC}
        \label{fig:RKSIaTS-TCC}
    \end{subfigure}\hfill
    \begin{subfigure}{0.18\textwidth}
        \centering
        \includegraphics[width=\linewidth]{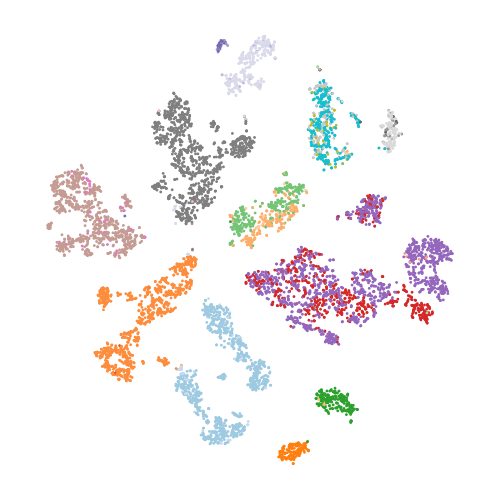}
        \caption{TS2Vec}
        \label{fig:RKSIaTS2Vec}
    \end{subfigure}\hfill
    \begin{subfigure}{0.18\textwidth}
        \centering
        \includegraphics[width=\linewidth]{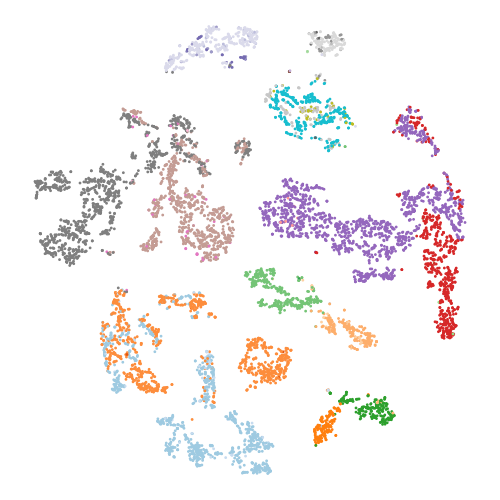}
        \caption{InfoTS}
        \label{fig:RKSIaInfoTS}
    \end{subfigure}\hfill
    \begin{subfigure}{0.18\textwidth}
        \centering
        \includegraphics[width=\linewidth]{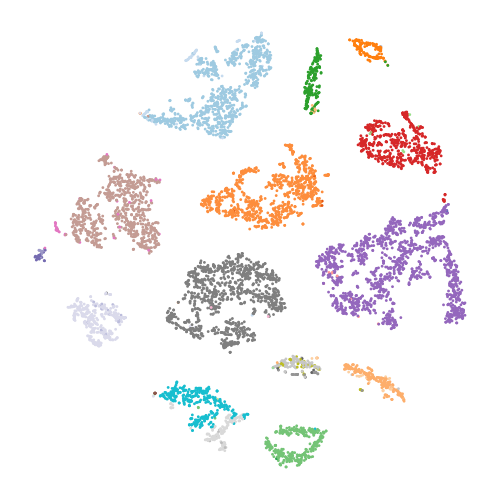}
        \caption{ATSCC}
        \label{fig:RKSIaATSCC}
    \end{subfigure}
    \hspace{0.09\textwidth}

    \caption{t-SNE Visualization of Learned Representations from the Arrival Trajectory Dataset at Incheon International Airport (RKSIa) Across All Representation Learning Baselines}
    \label{fig:tsne}
\end{figure*}

\subsubsection{Trajectory Classification}
An effective representation should have fidelity to class labels and enable linear separability. These properties can be evaluated using classification benchmarks, as instance-level labels \(Y\) were manually labeled according to AIPs. The evaluation follows the same standard classification protocol as demonstrated in \cite{TLoss, TS2Vec}. This protocol involves training a Support Vector Machine (SVM) classifier with a Radial Basis Function (RBF) kernel on the learned representation of the training data. For TNC, we fitted an SVM using the Global Alignment Kernel (GAK) \cite{ICML2011Cuturi_489GAK}. We then assess the SVM's accuracy using the representations derived from the testing data. We have integrated the classification challenges in \cite{Bosson2018, Murca2016, Murca2018, Deng2022, Campfens2020Classification}, classifying both runways or approach patterns.

According to the accuracy results in Table \ref{tab:comparison}, we observed a significant improvement in the accuracy of the arrival dataset for Incheon Airport, which features a complex configuration with four parallel runways. The experiment on departure trajectory data showed slightly better accuracy, as departure trajectories are relatively straightforward compared to arrivals. At Stockholm Arlanda Airport, which features two parallel runways, the results follow the same trend as Incheon arrivals, resulting in the highest accuracy on this larger dataset. Lastly, Zurich Airport has the most distinguishable pattern, yet we still observed significantly better performance compared to the other baselines.

The ATSCC model outperforms all baselines in accuracy, demonstrating that its learned representation is class-preserving and easily separable in the encoding space. This leads to significant improvements in the classifier's ability to identify runways and terminal maneuvering procedures compared to state-of-the-art baselines, highlighting ATSCC's suitability for air traffic trajectory data. Fig \ref{fig:tsne} further illustrates the separability of the learned representation, with classes being well separated.

\subsubsection{Trajectory Clustering}
We evaluate the uniformity of the learned representation in the encoding space to assess whether a simple clustering algorithm, when employed on the representation, can effectively differentiate data instances and ensure correct clustering without relying on true labels. We conduct the clustering experiment similarly to \cite{TNC} by implementing K-Means on the instance-level representation of trajectories using Euclidean distance, and setting the number of clusters equal to the number of unique test labels; however, for TNC, DTW distance was used.

Due to the absence of instance labels in recent ATM clustering works (e.g., \cite{Olive2020DeepTrajectory, Corrado2020WeightedEuclidean, Basora2017, Chakrabarti2023Modeling, Eerland2016}), clusters are typically analyzed visually. Besides, metrics such as the Silhouette score, Calinski-Harabasz Index, and Davies-Bouldin Index (DBI) were used (e.g., \cite{ZengDAE2021, Chu2022EnsembleClustering, Liu2022DeepTrajectory, Fan2023TALSTM, Damanik2022}). However, the fidelity and accuracy of cluster assignments are more crucial than the proximity of representations within clusters. Therefore, following recent clustering literature \cite{xie2016DEC, Caron_2018_ECCV_DeepCluster, 9533714Concencluster, vardakas2024deepSoftSilhouetteScore}, we use the Normalized Mutual Information (NMI) score and Adjusted Rand Index (ARI) to evaluate the similarity of clustering results with the true labels. This evaluation method provides a sensible performance assessment by testing whether the learned representation enhances clustering to recognize unique maneuvering procedures and correctly assign labels to the samples.

The NMI and ARI results tabulated in Table \ref{tab:comparison} show that both scores are significantly enhanced across arrival datasets, from those with complex arrival trajectories, such as Incheon and Stockholm Arlanda with their parallel runways, to those with less ambiguous configurations, like Zurich. Although the other baselines perform well with departure data, our approach still achieves superior scores. These results validate that the learned representation aligns well with the operational semantics. For example, trajectories with landing patterns that are geometrically similar become distinguishable in the encoding space. The superior scores reflect the framework's capability to generate the representations with exceptional uniformity, enabling K-Means to establish high-fidelity clusters. The experiment further confirms that the ATSCC framework is a more suitable representation learning method for air traffic management trajectory data. Moreover, besides separability, uniformity can also be observed in Fig \ref{fig:tsne}, as the same classes align well in distinct groups.

We further conducted a clustering experiment using a greater number of clusters than the number of classes because trajectory clustering is typically used for trajectory categorization, which requires more clusters than procedures, according to \cite{Chu2022EnsembleClustering, Deng2022}. This experiment aimed to assess the ability of the representation to enable clustering to generate high-fidelity subclusters, as ideally, these clusters should be subclasses of the procedures. To maximize fidelity, it is important to maximize mutual information \cite{InfoTS, Tian2020Infomin}. Therefore, we evaluate the mutual information score (MI) on clusters generated by K-means, with the number of clusters ranging from the number of existing classes to 100 clusters.

\begin{figure}
  \centering
  \includegraphics[width=\linewidth]{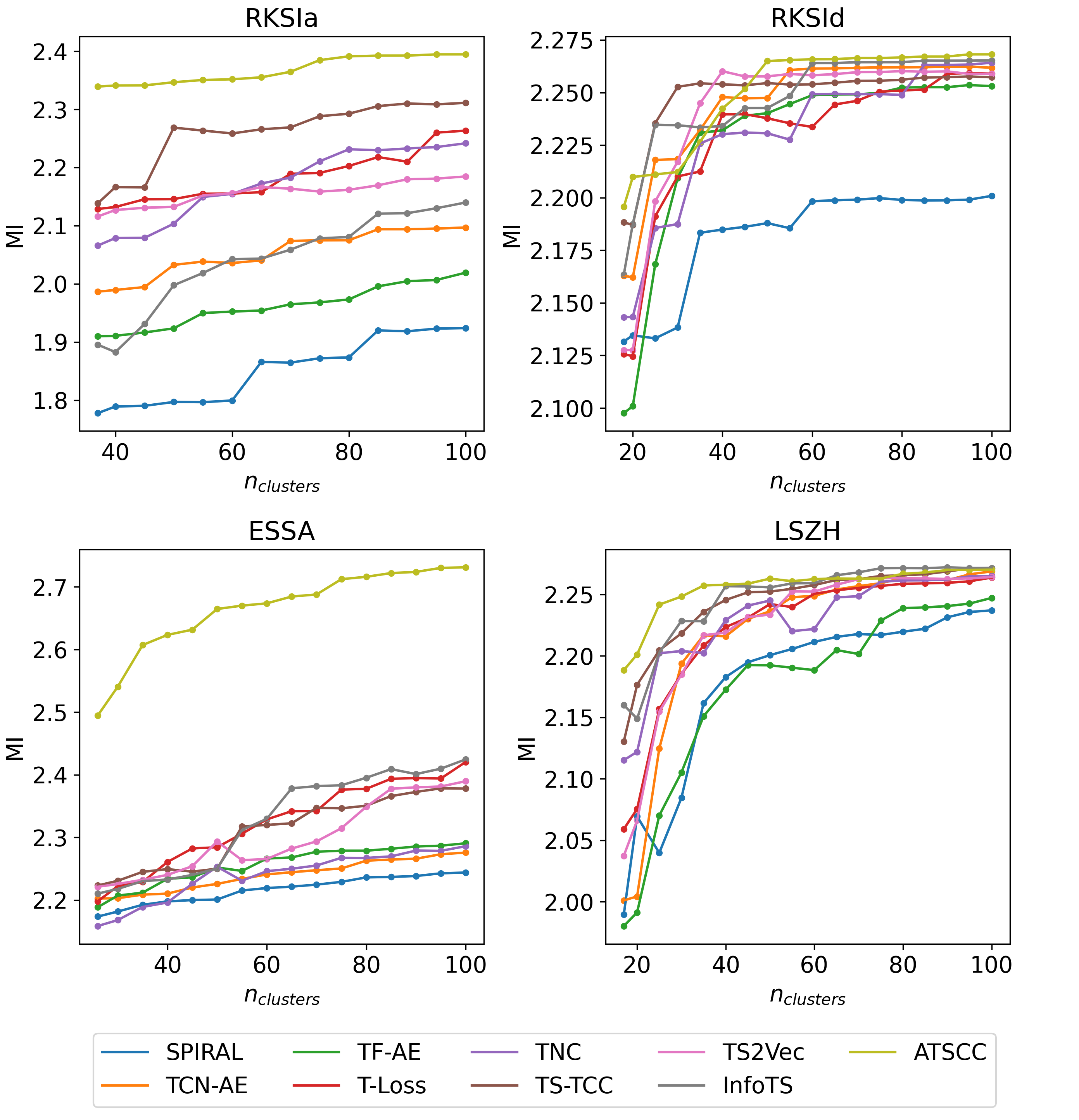}
  \caption{Comparison of MI Score Plot Evaluation on Learned Representation Using K-Means Clustering: Number of Clusters Ranges from the Number of Classes Up to 100, Incremented in Steps of 5}
  \label{fig:MI}
\end{figure}

Fig \ref{fig:MI} shows that our learned representation outperforms other methods for complex airports like Incheon and Stockholm Arlanda. However, for simpler cases such as Zurich and the departure trajectory of Incheon airports, ATSCC demonstrates slightly better performance due to minimal runway ambiguity. The MI scores converge to higher values, highlighting the importance of representation learning over the choice of clustering method and parameters. The results demonstrate that increasing the number of clusters to separate ambiguous groups is ineffective without an effective data representation. Along with classification and clustering benchmarks, this further highlights that the separability and uniformity of clusters lead to higher fidelity and accuracy in downstream tasks.

\subsection{Analysis}
\subsubsection{Ablation Study}
In this section, we evaluate the effectiveness of the framework's components by evaluating the average performance on four trajectory datasets upon modifications, as tabulated in Table \ref{tab:ablation}. We begin by analyzing the geometric feature extraction by excluding certain features and examining the impact on the results. Polar positions have a greater impact on performance than path vector sequences, as they change rapidly near the airport, resulting in distinct tokens. Conversely, adding path sequences as the only additional feature reduces scores due to potential confusion from parallel landing paths. However, we found that including both polar positions and path sequence significantly enhances overall performance.

We have observed that applying random binomial masking on the input and attention masks generates significantly effective representation because the random masking enhances regularization and generalization, performing similarly to dropout. Additionally, it provides more variation of data, similar to data augmentation.

L2 normalization, which constrains the representation within a hypersphere, can enhance downstream task performances by promoting linear separation in the encoding space. Therefore, we observed that eliminating L2 normalization after both projection layers resulted in significant performance loss, similar to L2 normalizing only on tokens. On the other hand, L2 normalization on the representation significantly enhanced the scores, consistent with \cite{Hemisphere}; however, slight additional improvement was observed when adding token normalization.

Given that the ATSCC encoder does not incorporate explicit positional encoding, we compared our results with those of commonly used positional encodings. We referenced \cite{Vaswani2017Transformer} for sinusoidal positional encoding and \cite{TST} for initializing the learnable positional encoding. We observed that the encoder following the NoPos \cite{NoPos} outperforms those with explicit positional encoding, which can alter the information in tokens. For self-supervised learning, it is important to utilize underlying data information, rather than relying solely on expected output as in supervised learning. Introducing distortions through positional encoding can fundamentally impact training.

We verify the suitability of the casual transformer by implementing different encoder backbones. The ATSCC performs binomial masking on both the input and the attention matrices. However, different architectures perform this random masking differently. The dilated causal convolution network encoder architecture and its random masking technique are referenced in \cite{TS2Vec}. Later, we replace the backbone of this CNN model with a 3-layer LSTM model having a hidden layer dimension of 512. Both encoders lead to a significant decrease in overall performance, as CNN can become locally focused, and LSTM loses information when encoding long sequences.

The soft nearest neighbor loss \cite{frosst2019analyzingSNNL} was rearranged and modified into Equation (\ref{eq:snnlnew}). We have empirically proven that omitting the positive contrasting terms in the sum with negatives results in better performance in our framework and experiment setting.

\begin{table}
\caption{Ablation Results on Four Trajectory Datasets}
\centering
\begin{tblr}{
  cell{1}{2-4} = {c=1}{c},
  hline{1-3,7,9,13,16,19,21} = {-}{},
}
Configurations                                                                                             & ACC    & NMI    & ARI    \\
Full Configuration                                                                                         & 0.9977 & 0.8993 & 0.7868 \\
Geometric Feature Extraction                                                                               &        &        &        \\
\labelitemi\hspace{\dimexpr\labelsep+0.5\tabcolsep}without aircraft path vectors                          & 0.9971 & 0.8751 & 0.7537 \\
\labelitemi\hspace{\dimexpr\labelsep+0.5\tabcolsep}without polar positions                                 & 0.9966 & 0.7883 & 0.5624 \\
\labelitemi\hspace{\dimexpr\labelsep+0.5\tabcolsep}catesian positions only                                 & 0.9974 & 0.8080 & 0.6588 \\
Random Masking                                                                                             &        &        &        \\
\labelitemi\hspace{\dimexpr\labelsep+0.5\tabcolsep}without random masking                                  & 0.9962 & 0.8044 & 0.5818 \\
L2 Normalization                                                                                           &        &        &        \\
\labelitemi\hspace{\dimexpr\labelsep+0.5\tabcolsep}without token L2 norm                                   & 0.9977 & 0.8941 & 0.7691 \\
\labelitemi\hspace{\dimexpr\labelsep+0.5\tabcolsep}without representation L2 norm                          & 0.9952 & 0.7902 & 0.5784 \\
\labelitemi\hspace{\dimexpr\labelsep+0.5\tabcolsep}without both L2 norm                                    & 0.9961 & 0.8090 & 0.6089 \\
Positional Encoding                                                                                        &        &        &        \\
\labelitemi\hspace{\dimexpr\labelsep+0.5\tabcolsep}Sinusoidal                                              & 0.9723 & 0.1863 & 0.0712 \\
\labelitemi\hspace{\dimexpr\labelsep+0.5\tabcolsep}Learnable                                               & 0.9892 & 0.7751 & 0.6400 \\
Backbone Architecture                                                                                      &        &        &        \\
\labelitemi\hspace{\dimexpr\labelsep+0.5\tabcolsep}LSTM                                                    & 0.6195 & 0.5359 & 0.2606 \\
\labelitemi\hspace{\dimexpr\labelsep+0.5\tabcolsep}Dilated Causal Convolution                              & 0.9456 & 0.4903 & 0.2752 \\
soft nearest neighbor loss                                                                                 &        &        &        \\
\labelitemi\hspace{\dimexpr\labelsep+0.5\tabcolsep} $\mathcal{L}_{snn}$: Equation (\ref{eq:snnlrearrange}) & 0.9973 & 0.8718 & 0.7361
\end{tblr}
\label{tab:ablation}
\end{table}

\begin{figure}[htbp] 
    \centering
    \begin{subfigure}{\linewidth} 
        \includegraphics[width=\linewidth]{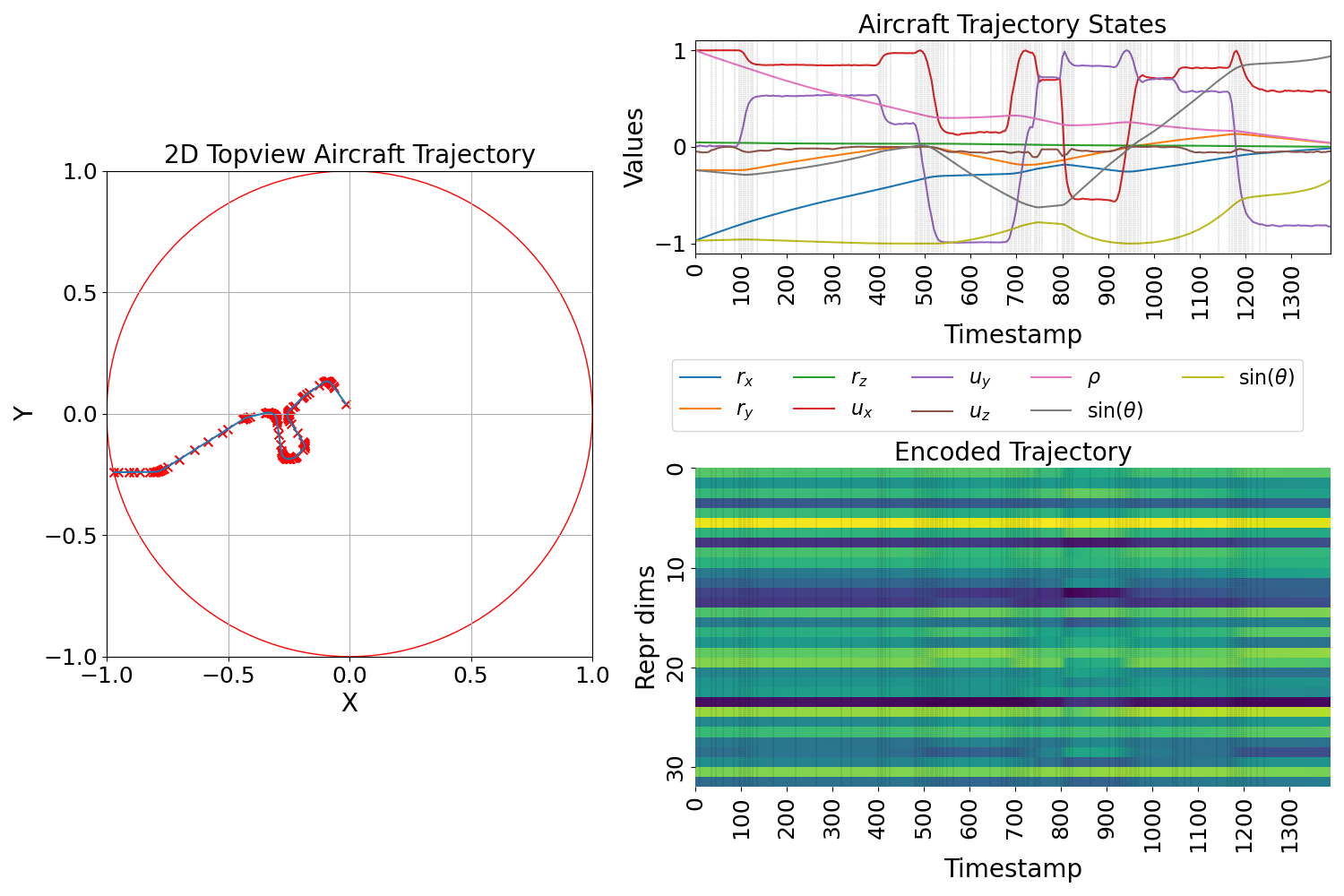}
        \caption{Trajectory of Class 8: Arriving via STAR REBIT2, Landing on Runway 15L via IAF MUNAN}
        \label{fig:spiral1} 
    \end{subfigure}
    \begin{subfigure}{\linewidth} 
        \includegraphics[width=\linewidth]{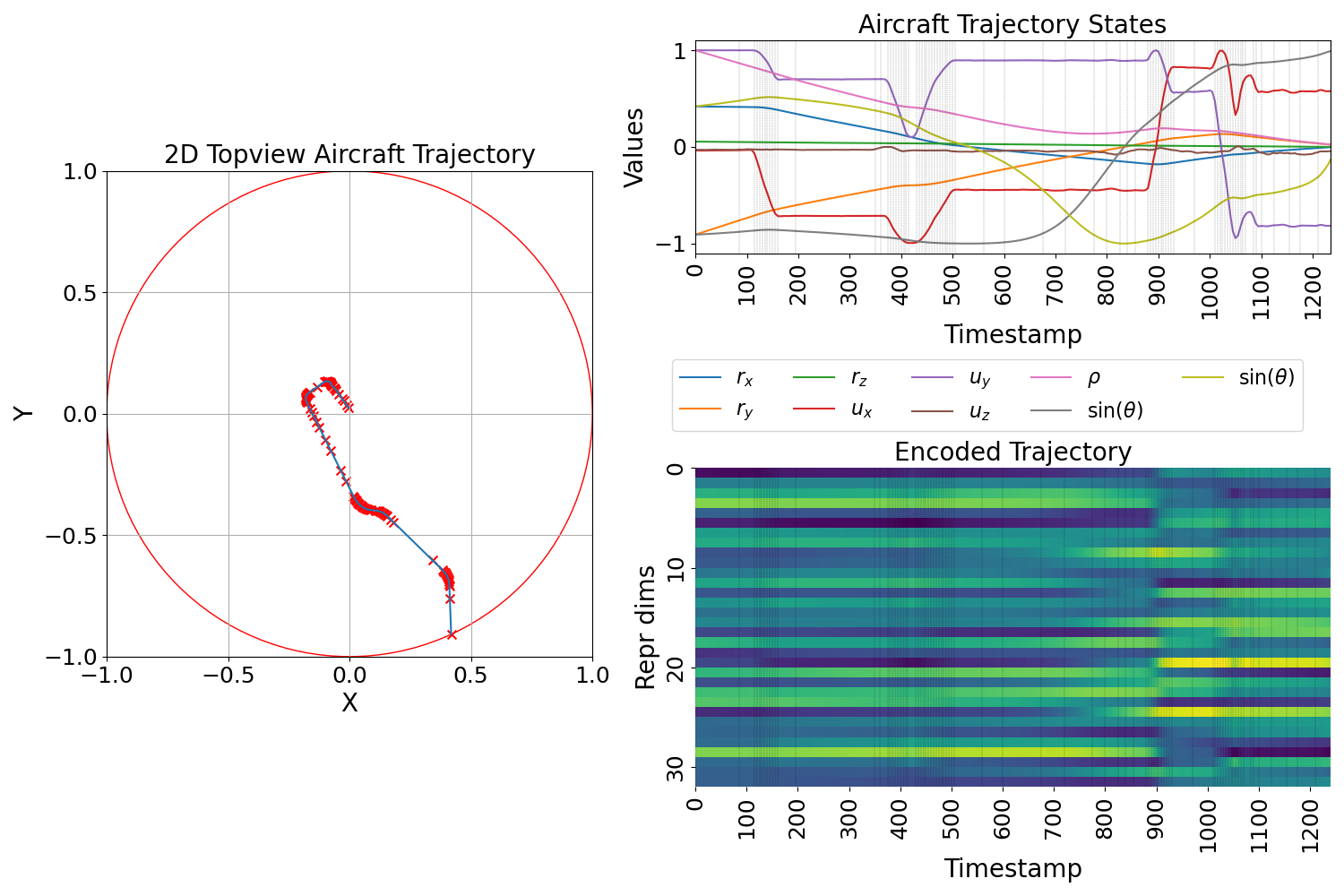} 
        \caption{Trajectory of Class 28: Arriving via STAR OLMEN2H, Landing on Runway 15L via IAF MUNAN}
        \label{fig:spiral2} 
    \end{subfigure}
    \caption{Visualization of Embedded Trajectories, 2D Top-View Trajectory Plot and Geometric Features} 
\label{fig:embedding}
\end{figure}

\begin{figure*}[htbp]
  \centering
  \includegraphics[width=0.8\textwidth]{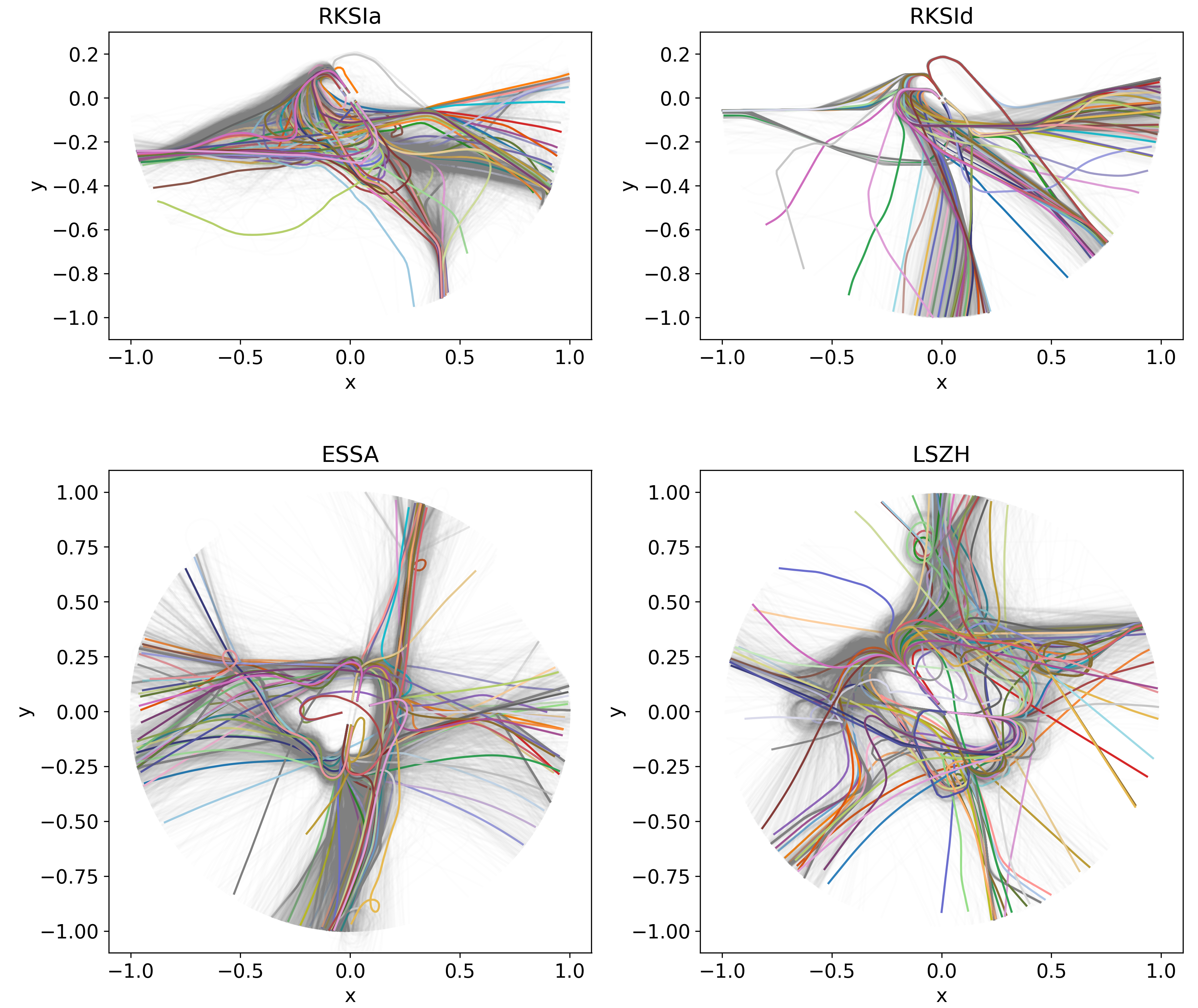}  
  \caption{Visualization of Clustering Results on Four Trajectory Datasets: Illustrating Representative Trajectories Using the Mean of Cluster Members}
  \label{fig:clus}
\end{figure*}

\subsubsection{Visual Explanation of Embedding}
For ease of visual explanation, we have taken two trajectories from the Incheon arrival dataset and their learned representations at all timestamps, which were trained with the output dimension set at 32. Both flights landed on the same runway but arrived via different STARs. At each timestamp, trajectory states in different scales and units are collectively converted into scale-invariant representation vectors. According to Fig \ref{fig:embedding}, the representation $z_{i,t}$ changes over time as causal attention collects information $x_{i,t}$ in $\mathcal{X}'_{i,t} \subseteq \mathcal{X}_i$ from current and previous states at each timestamp $t \in \{1, \dots, t\}$. A clear difference in \( z_{i,t} \) of the two trajectories is noticeable at earlier timestamps due to different STARs. Later, \( z_{i,t} \) of two trajectories diverges, despite landing on the same runway, because the information of the STARs was carried along while encoding. The ATSCC does not forcibly discretize the states, nor does it use the notation of Euclidean vectors or waypoints. Instead, the values $z_{i,t}$ represent scale-invariant semantic embeddings that preserve temporal transitions in continuous encoding space. This enables their applicability to machine learning models, which can be flexibly adapted for other tasks.

\subsubsection{Analysis for Trajectory Categorization}
Many previous studies have focused on air traffic categorization; therefore, this section shows the feasibility of our model in addressing this problem. The trajectory can be analyzed by varying the number of clusters to 100 or more, as suggested by \cite{Liu2023, Deng2022}. Fig. \ref{fig:MI} shows that increasing the number of clusters maximizes the MI score; thus, we performed a brief analysis using 100 clusters as an example of clustering results for a categorization task.

Since the trajectory data undergoes semantic analysis using the learned representation, trajectories with spatially close but operationally distinct segments, such as parallel landing paths, can be differentiated, as demonstrated by the results on Incheon and Stockholm Arlanda Airports (Fig. \ref{fig:clus}). Unlike other studies that address complex airports with parallel runways (e.g., \cite{ZengDAE2021, Chu2022EnsembleClustering, Murca2016, Murca2018, Deng2022, Andrienko2013, Damanik2022}), ATSCC effectively enables separation of ambiguous classes without requiring airport information, resulting in high-fidelity subclasses of the procedures. Similar to \cite{Liu2023}, the visualization shows the identifications of non-standard patterns, notably at Incheon Airport, and trajectories with holding patterns are clearly grouped at both Stockholm Arlanda and Zurich Airport. This confirms that the ATSCC is also applicable for characterizing trajectories and outlier detection.

\section{CONCLUSION}
This paper has proposed a representation learning framework for multivariate air traffic trajectory data. ATSCC trains a causal transformer encoder by assigning segment IDs via the RDP algorithm for training with the soft nearest neighbor loss. We proposed fidelity benchmarking using manually labeled trajectory data, guided by AIPs, to evaluate the representation. The classification and clustering results show that ATSCC is the most suitable framework for ATM trajectory data. Our learned representation enables the separation of ambiguous paths that are semantically different, non-standard patterns, and holding points. We have addressed the limitations of several prior works, for ATSCC does not rely on computationally intensive distance metrics, nor does it require the airport information. Patterns are recognized through the training, showing adaptability to various airport configurations. The recommendations for future works include extending the model's applicability to other trajectory or time series data or adapting the model to tasks such as trajectory prediction, imputation, or anomaly detection. Moreover, ATSCC can potentially train a pre-trained trajectory model for various machine-learning tasks. Additionally, incorporating weather or operational data could extend the model’s utility across diverse scenarios.


\bibliographystyle{IEEEtran}
\bibliography{IEEEbib}

\begin{IEEEbiography}[{\includegraphics[width=1in,height=1.25in,clip,keepaspectratio]{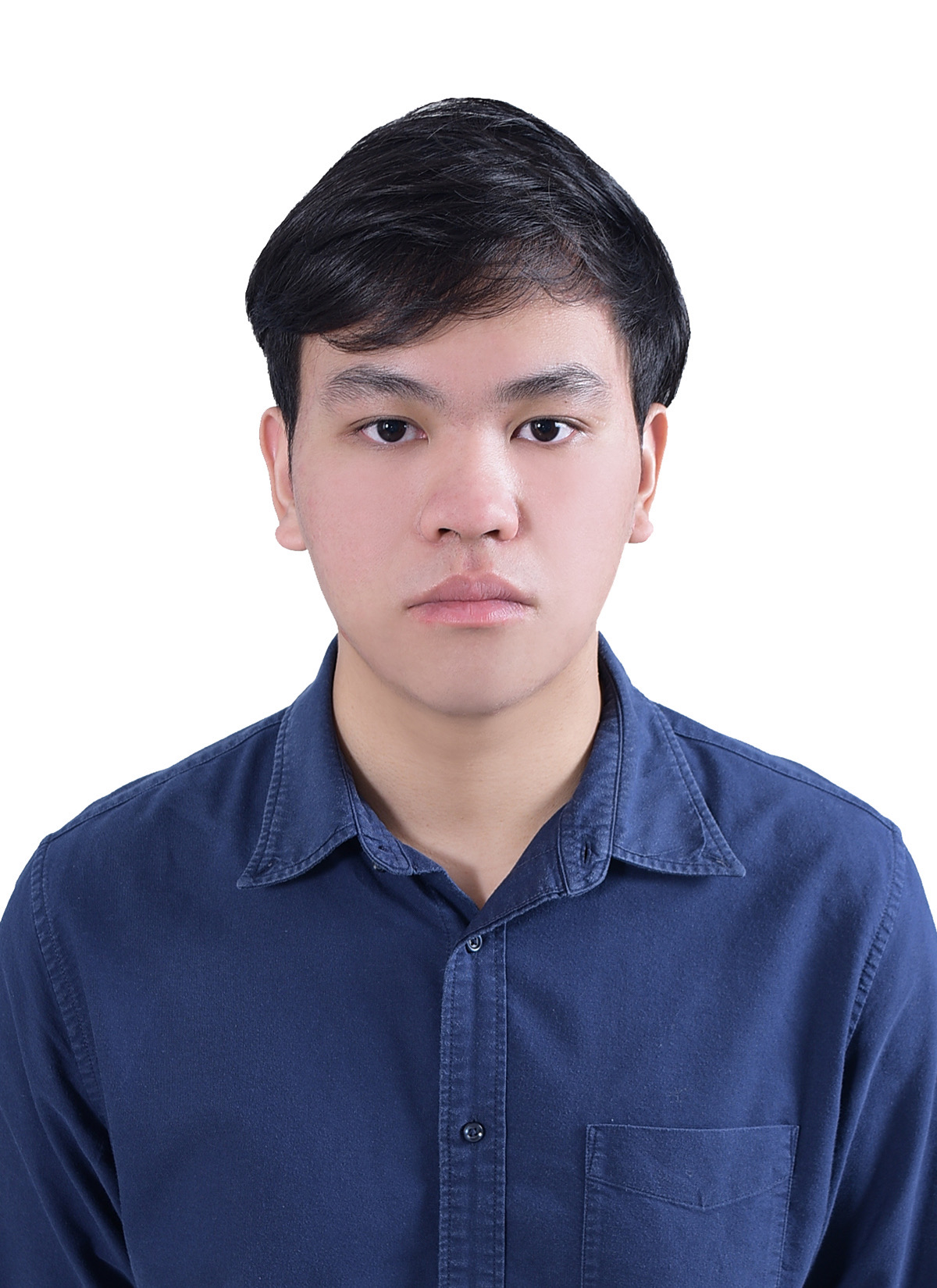}}]{Thaweerath Phisannupawong} received a B.Eng. degree in Aeronautical Engineering and Commercial Pilot from King Mongkut's Institute of Technology Ladkrabang (KMITL), Bangkok, Thailand, in 2021. Since 2022, he has been pursuing an M.S. degree in Aerospace Engineering at the Korea Advanced Institute of Science and Technology (KAIST) in Daejeon, South Korea. His research focuses on aerospace data applications and representation learning.
\end{IEEEbiography}

\begin{IEEEbiography}[{\includegraphics[width=1in,height=1.25in,clip,keepaspectratio]{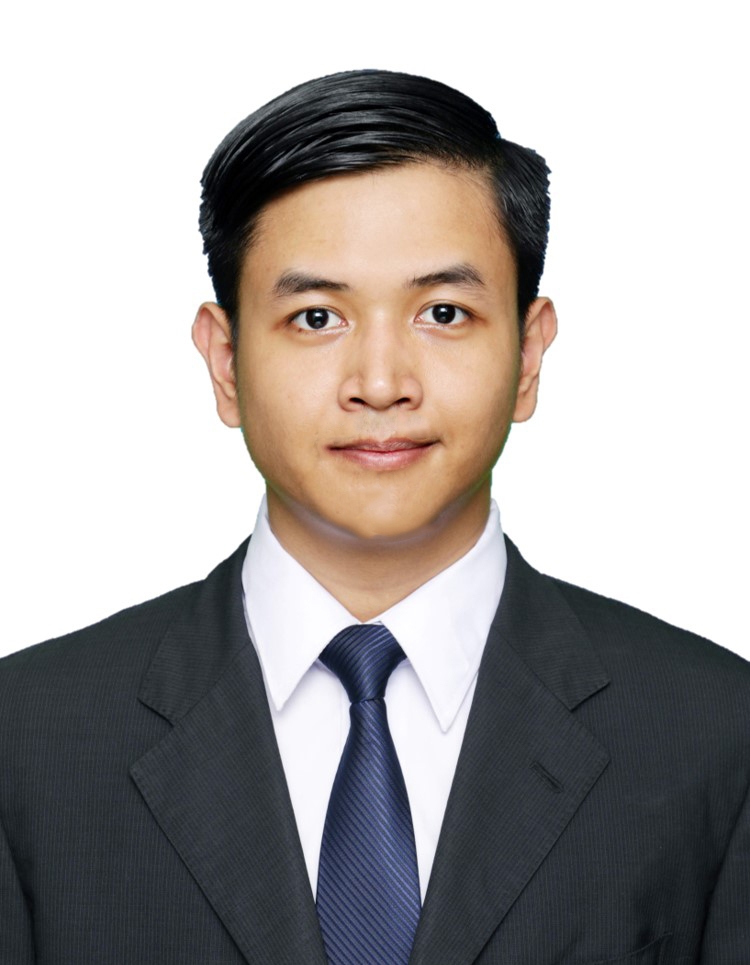}}]{Joshua J. Damanik} received the B.S. degree in engineering physics from Institut Teknologi Bandung, Indonesia, in 2018, and the M.S. degree in aerospace engineering from the Korea Advanced Institute of Science and Technology (KAIST), Daejeon, South Korea, in 2021, where he is currently pursuing the Ph.D. degree in aerospace engineering. His current research interests include robotics estimation and control, and data mining.
\end{IEEEbiography}

\begin{IEEEbiography}[{\includegraphics[width=1in,height=1.25in,clip,keepaspectratio]{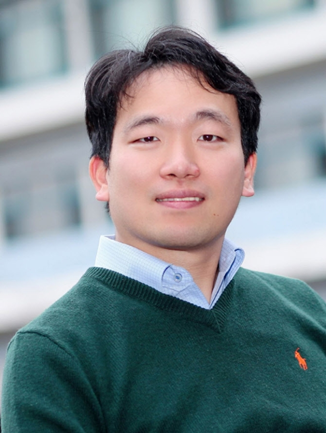}}]{Han-Lim Choi} (Senior Member, IEEE) received the B.S. and M.S. degrees in aerospace engineering from the Korea Advanced Institute of Science and Technology (KAIST), Daejeon, South Korea, in 2000 and 2002, respectively, and the Ph.D. degree in aeronautics and astronautics from the Massachusetts Institute of Technology (MIT), Cambridge, MA, USA, in 2009. Then, he studied at MIT as a Postdoctoral Associate until he joined KAIST, in 2010. He is currently a Professor of aerospace engineering at KAIST. His research interests include estimation and control for sensor networks and decision making for multi-agent systems. He was a recipient of the Automatic Applications Prize, in 2011 (together with Dr. Jonathan P. How).
\end{IEEEbiography}

\end{document}